%% file: iclr2027_conference.tex
\documentclass{article} 
\usepackage{iclr2027_conference,times}

\input{math_commands.tex}

\usepackage[utf8]{inputenc} 
\usepackage[T1]{fontenc}    
\usepackage{hyperref}       
\usepackage{url}            
\usepackage{booktabs}       
\usepackage{amsfonts}       
\usepackage{nicefrac}       
\usepackage{microtype}      
\usepackage{xcolor}         

\usepackage{wrapfig}
\usepackage{multirow}
\usepackage{amsmath, amsthm}
\usepackage{amssymb}
\usepackage{algorithm}
\usepackage{algpseudocode}
\usepackage{graphicx}
\usepackage{float}
\usepackage{siunitx}
\usepackage[table]{xcolor}  
\usepackage{array}
\usepackage{stackengine}
\definecolor{diffblue}{HTML}{0000FF}
\newcolumntype{H}{>{\columncolor{blue!10}}c}

\definecolor{firstcolor}{RGB}{243, 232, 253}  
\definecolor{secondcolor}{RGB}{218, 232, 252} 
\definecolor{thirdcolor}{RGB}{234, 234, 234} 

\definecolor{kwtabg}{RGB}{243, 232, 253} 

\newtheorem{definition}{Definition}

\title{SpikeMoE: Brain-Inspired Competitive Routing for Flexible Spiking Mixture-of-Experts}

\author{Xiaoli Liu \quad Yujie Liang \quad Jialin Li \quad Chenxiao Dou \quad Malu Zhang \\
School of Computer Science and Engineering,\\
University of Electronic Science and Technology of China}

\iclrfinalcopy 
\begin{document}

\maketitle
\lhead{Preprint}

\begin{abstract}
Spiking Neural Networks (SNNs) enable event-driven computation through biologically inspired dynamics at the neuronal scale, while Mixture-of-Experts (MoE) perform conditional computation through expert selection at the model scale. Integrating their strengths offers potential for flexible neural architectures. A key challenge, however, lies in designing an expert selection mechanism based on spiking activity. To address this, we introduce a spike-based k-WTA Router inspired by competition-inhibition observed in the hippocampal CA1 region. The router incorporates lateral inhibition and refractory period to select Top-K experts according to discrete spike counts. Building on this, we present SpikeMoE, a framework that integrates neuronal-scale spiking dynamics with model-scale expert selection. To address incomplete multisensory inputs in multimodal tasks, we further equip SpikeMoE with a two-stage missing-modality modeling module that combines empirical prototypes from an observed-modality pool with modality-specific learnable embeddings to construct missing-modality representations. Experiments on vision, language, and multimodal benchmarks demonstrate that SpikeMoE achieves state-of-the-art performance among the SNN baselines, matches or exceeds the performance of ANN counterparts, and maintains robustness across diverse missing-modality conditions. These results demonstrate a favorable trade-off between performance and energy efficiency, validating the integration of spiking dynamics with sparse expert computation and highlighting SpikeMoE as a promising approach to energy-efficient brain-inspired computing.
\end{abstract}

\section{Introduction}

The rapid advancement of artificial intelligence has given rise to increasingly complex neural network models, achieving success in numerous fields. However, modern neural networks consume vast amounts of computation and memory \citep{hwang2024spikedattention}. With the continuous growth in demand for intelligent systems, developing energy-efficient computing paradigms has become a critical priority. Nature provides an encouraging blueprint for high-efficiency computing. The human brain executes complex cognitive tasks with far greater efficiency than modern AI systems, all while consuming significantly less energy \citep{amunts2016human}. This remarkable efficiency stems from two key principles: at the microscopic level, biological neurons are activated by discrete, event-driven spikes rather than continuous signals \citep{roy2019towards}, enabling sparse and asynchronous computation \citep{yao2023spike,yaospike}; at the macroscopic level, the brain exhibits distributed processing by selectively activating distinct regions for specific tasks \citep{o2021and}, enabling robust environmental perception through precisely coordinated multisensory integration.

\begin{figure*}[htbp]
    \centering    
    \includegraphics[width=1.0\textwidth]{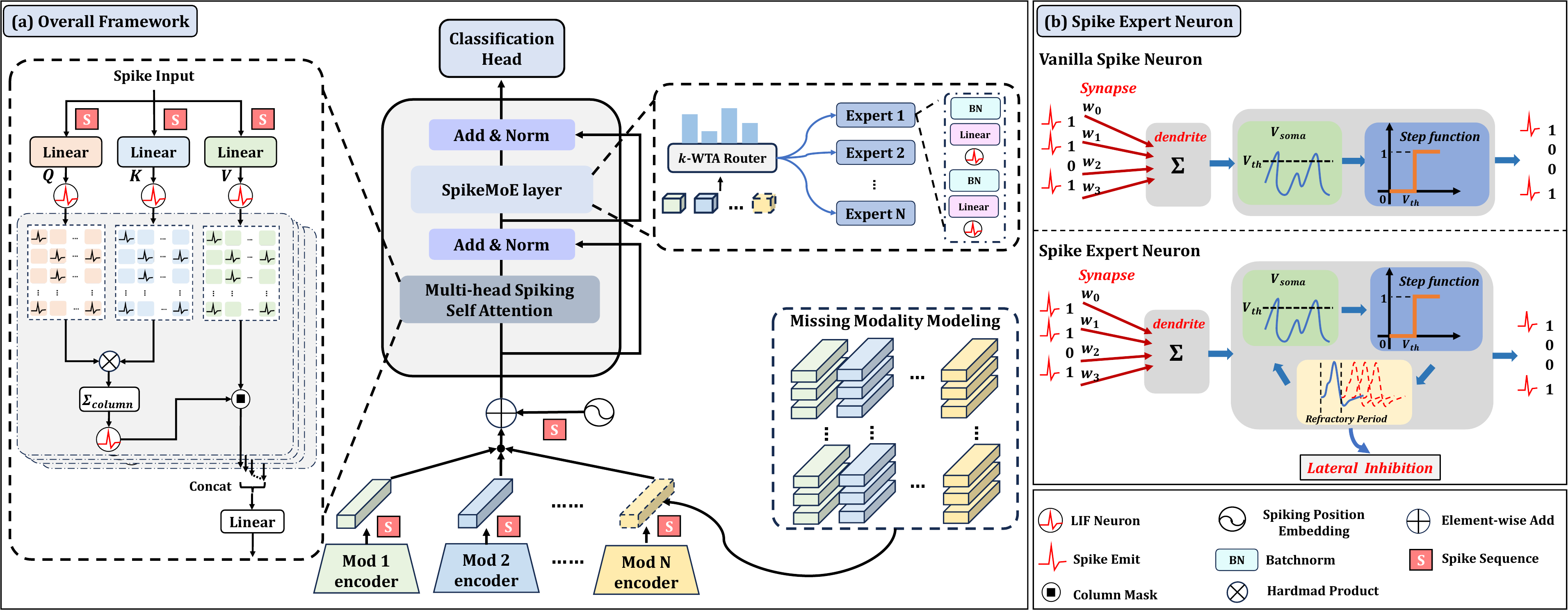}  
    \vspace{-5mm}
    \caption{(a) Overall general framework; (b) Spike expert neuron.}
    \vspace{-5mm}
    \label{fig:main figure}
\end{figure*}

Spiking Neural Networks (SNNs) \citep{izhikevich2003simple,maass1997networks,masquelier2008spike}, considered the third generation of neural networks \citep{maass1997networks}, simulate the dynamics of biological neurons at the microscopic level \citep{roy2019towards}. The event-driven \citep{wei2024event} characteristic of SNNs means that neurons fire spikes only when activated and remain silent otherwise \citep{akopyan2015truenorth}. This mechanism increases the sparsity of synaptic computation and eliminates multiplication-accumulation (MAC) operations \citep{davies2018loihi,zhang202322, wang2023spatial}, thereby significantly improving energy efficiency. Furthermore, Mixture-of-Experts (MoE) \citep{mustafa2022multimodal} achieve dynamic expert selection through conditional computation \citep{shazeer2017outrageously, jacobs1991adaptive, chen1999improved, yuksel2012twenty}, which aligns with the distributed and sparse processing at the macroscopic brain-region level \citep{o2021and}.


Combining SNNs with MoE is promising because they provide complementary forms of efficient computation: SNNs exploit sparse, event-driven neuronal dynamics, whereas MoE performs conditional computation through expert selection. Nevertheless, developing a flexible spiking MoE presents two related challenges. At the mechanism level, conventional MoE routers typically generate continuous expert scores using softmax, whose exponentiation and global normalization operations are not naturally compatible with the discrete, event-driven computation of SNNs. A central challenge is therefore to perform effective expert selection directly based on spiking activity. At the perception level, multisensory inputs in real-world scenarios are frequently incomplete \citep{han2024fusemoe,glasser2016multi,xu2023multimodal,wu2024deep,yuan2025survey,zhang2022mmformer}. Missing modalities deprive the router and experts of important contextual cues, potentially disrupting expert selection. Without appropriately representing the missing information and exploiting the available modalities, the resulting decisions can become fragile and suboptimal \citep{shaik2024survey,cao2023multi,tsai2019multimodal,jiang2023cmci,peng2022balanced}.


To address these challenges, we propose SpikeMoE, a unified and flexible framework that integrates the strengths of SNNs and MoE. To resolve the internal bottleneck, we draw inspiration from hippocampal CA1 calcium imaging to propose k-WTA Router which is based on competition-inhibition mechanism \citep{topolnik2022role}. This approach achieves dynamic expert selection via sparse spike activation, effectively mitigating the issue of expert collapse. To overcome external perception challenges and improve robustness against incomplete inputs, we introduce a two-stage missing-modality modeling strategy. This enables the model to flexibly handle arbitrary combinations of modalities. The main contributions are summarized as follows:


\begin{itemize}
    \item We introduce a k-WTA Router based on competition-inhibition mechanism. By incorporating lateral inhibition and refractory dynamics, the router performs Top-K expert selection according to spike counts and mitigates repeated within-input activation of a single expert.
    \item We develop a two-stage missing-modality modeling strategy that combines empirical information from an modality pool with modality-specific learnable embeddings, which enables the robust modeling of arbitrary modality combinations.
    \item By integrating the k-WTA Router with the missing-modality modaling strategy, we present SpikeMoE, a unified and flexible framework that connects neuronal-scale spiking dynamics with modal-scale sparse expert activation.
    \item Extensive evaluations show that SpikeMoE achieves performance comparable to or better than its ANN counterparts and establishes state-of-the-art results among the evaluated SNN baselines. Moreover, it remains robust under diverse missing-modality conditions while offering a favorable performance–energy efficiency trade-off.
\end{itemize}

\section{Related Works}

\textbf{Multimodal learning in SNNs.} With recent advances in neuromorphic computing, exploration of multimodal SNNs has gained momentum. We summarize main studies according to different categories. \textbf{a. Audiovisual classification.} In \citep{liu2025towards}, the introduction of MLIF neurons synchronizes and coordinates the spike activation of audiovisual signals within a single neuron; Liu et al. \citep{liu2022event} proposed an event-based multimodal SNN comprising visual and auditory unimodal subnetworks and an attention-based cross-modal subnetwork for fusion. A spiking multimodal transformer was also proposed for multimodal audiovisual classification \citep{guo2023transformer}. \textbf{b. Emotion Recognition.} Tan et al. \citep{tan2020fusionsense} employed an evolutionary architecture of multimodal SNNs based on the NeuCube framework; the Sliding Parallel Spiking Conv-ViT \citep{chen2025enhancing} integrated facial expressions, speech, and text, utilizing aligned heatmap features and multiscale wavelet transforms for precise feature extraction.

\textbf{Multimodal learning in missing modality scenario.} In many tasks, considering interactions among multimodal data can provide comprehensive understanding for downstream applications. MISTS \citep{zhang2022mmformer} employs a multi-time attention mechanism to encode irregular time series and utilizes multimodal cross-attention to fuse different modalities. In real-world scenarios, missing modalities pose a critical challenge. \citep{du2018semi,shang2017vigan} introduce generative adversarial networks to estimate missing modalities through semi-supervised learning; ShaSpec \citep{wang2023multi} employs multiple loss functions to learn the proportion and specific representations of different modalities. However, the reconstruction networks in ShaSpec require multiple additional auxiliary loss functions, consuming substantial computational resources during optimization. Furthermore, recent studies like FuseMoE \citep{han2024fusemoe} respect modal absence and aim to assign each combination of multimodal data to distinct experts.

\section{Preliminaries}
\subsection{Spiking Neuron Model}

SNNs achieve spike-driven processing using nonlinear spiking neurons. The Leaky Integrate-and-Fire (LIF) model is a widely adopted mathematical abstract for spike-driven behavior, achieving a balance between biological plausibility and computational efficiency \citep{maass1997networks}. The dynamic model of LIF neuron is described as:
\begin{align}
    & H[t]=V[t-1]+\frac{1}{\tau}(X[t]-(V[t-1]-V_{reset})),  \\
    & S[t]=\Theta(H[t]-V_{th}),  \\
    & V[t]=H[t](1-S[t])+V_{reset}S[t], 
\end{align}
where \(\tau\) is the membrane time constant, and \(X[t]\) is the input current at time step \(t\). When the membrane potential \(H[t]\) exceeds the firing threshold \(V_{th}\), the spike neuron will trigger a spike \(S[t]\). \(\Theta(\cdot)\) is the Heaviside step function which equals \(1\) for \(v \ge 0\) and \(0\) otherwise. \(V[t]\) represents the membrane potential after the trigger event which equals \(H[t]\) if no spike is generated, and otherwise equals to the reset potential \(V_{reset}\).

\subsection{Mixture-of-Experts Architecture}

The Mixture-of-Experts (MoE) \citep{shazeer2017outrageously, jacobs1991adaptive, chen1999improved, yuksel2012twenty} enables selective expert activation, where only the top-k experts process each token. The MoE layer consists of a collection of \(m\) experts, \(\textbf{E} = \{\textbf{E}_1, \textbf{E}_2, ..., \textbf{E}_m\}\), and a router \(\mathcal{R}_{\text{A}}\) that selects the appropriate experts from this collection. Given the input sequence \(\textbf{X}_{\text{A}}\), the output is formulated as the sum of the contributions of the top-k experts selected from the \(m\) candidates using the router:
\begin{align}
    \text{y}&=\sum^K_{k=1} \mathcal{R}_k(\textbf{X}_{\text{A}}) \cdot \textbf{E}_k(\textbf{X}_{\text{A}}), \\
    \mathcal{R}(\textbf{x})&=\text{TopK}(\text{softmax}(\textbf{X}_{\text{A}} \textbf{W}_{\mathcal{R}},K)), \\
    \text{TopK}(\textbf{v},K)&=
        \begin{cases}
            \textbf{v} \ \ \text{if} \ v \ \text{is in the top} \ K \ \text{elements} \\
            0 \ \ \text{otherwise}
        \end{cases},
\end{align}
where \(\textbf{W}_{\mathcal{R}}\) denotes the weight matrix of the router. \(\text{TopK}\) combined with \(\text{softmax}(\cdot)\) sets all elements of the routing vector to \(0\) except for the top K maximum values.

\section{Methods}
\label{methods}

\subsection{K-WTA Router in SpikeMoE}

Building upon the original Transformer architecture, MoE introduces a substantial number of parameters and achieves conditional computation by calculating the routing probability via softmax function. However, SNNs must avoid multiplication, floating-point routing experts and softmax involving exponentiation and division do not adhere to the computational principles of SNNs. Therefore, it is necessary to design SNN-friendly gated routing mechanisms.

\begin{wrapfigure}{r}{0.6\textwidth}
\vspace{-5mm}
\centering 
\includegraphics[width=\linewidth]{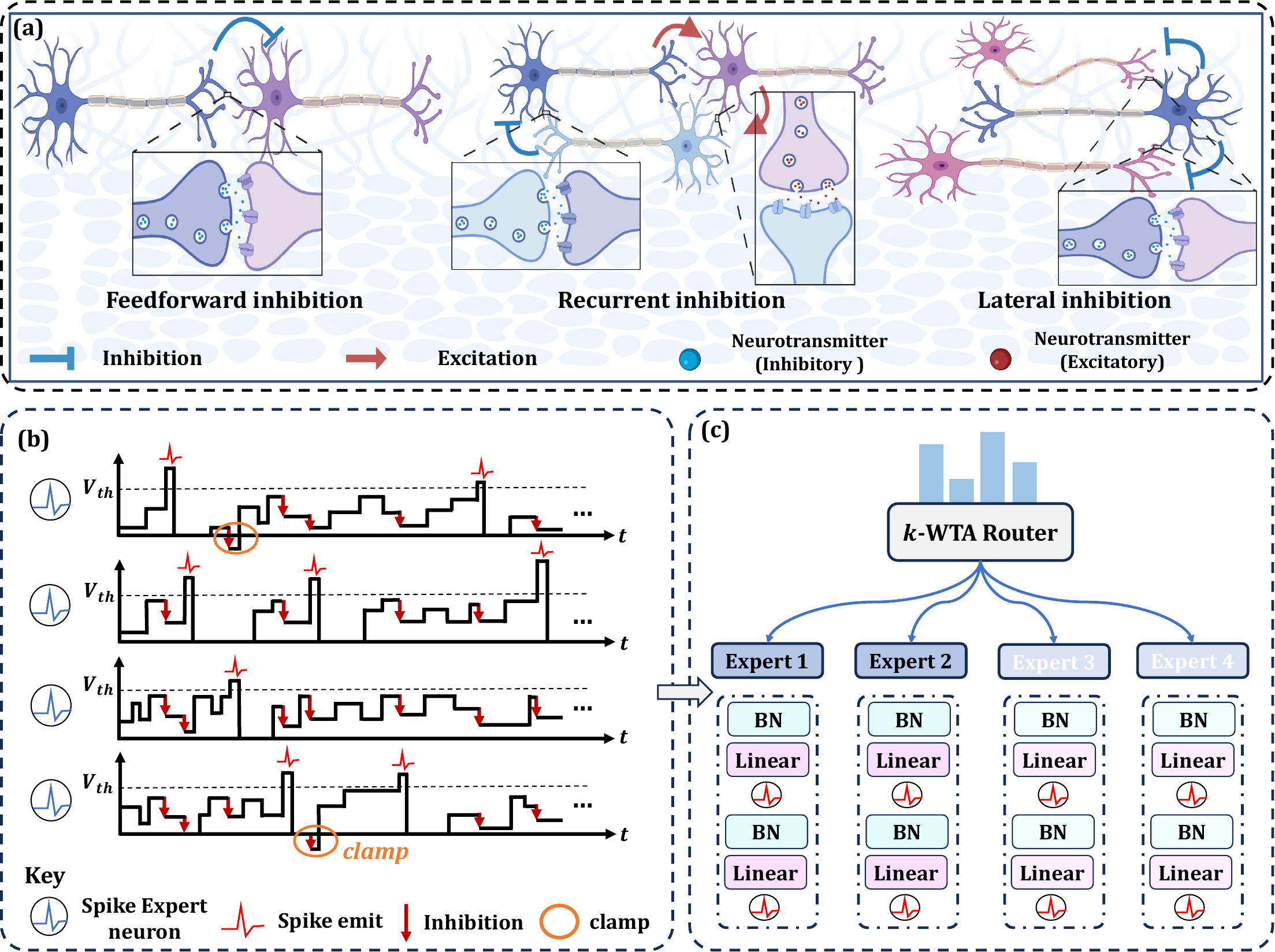}
\vspace{-6mm}
\caption{(a) Common inhibitory patterns in the hippocampus; (b) Competitive selection in k-WTA Router; (c) Expert selection.}
\vspace{-6mm}
\label{fig:k-WTA Router}
\end{wrapfigure}

\textbf{Competition-inhibition mechanism.} Neuroscientists have observed lateral inhibition in calcium imaging results from the hippocampal CA1 region (Figure \ref{fig:k-WTA Router}(a)) \citep{topolnik2022role}. It plays a vital role in information encoding and processing by enhancing neurons' specific responses to stimuli and improving spatial resolution in information processing \citep{topolnik2022role, jeong2022learning}. It serves as a crucial mechanism for selective learning. Inspired by this phenomenon, we propose a novel spiking gating network based on a competition-inhibition mechanism, named the k-WTA (Winner-Take-All) Router. Specifically, it's constructed based on an improved LIF neuron model (Figure \ref{fig:main figure}(b)), whose dynamics are described by the following discrete-time equation:
\begin{equation}
\label{equ:lif}
    V[t]=V[t-1] \cdot (1-\frac{1}{\tau})+I[t],
\end{equation}
where \(V[t] \in \mathbb{R}^{B\times E}\) denotes the membrane potential tensor at time step \(t\), \(B\) is the batch size, \(E\) is the number of experts, \(\tau\) is the time constant, and \(I[t]\) is the input current. To promote competitive selection among experts, we introduce a learnable lateral inhibition mechanism:
\begin{equation}
\label{equ:inhibit}
    V_{inhibit}[t]=V[t]-S[t-1] \cdot (L \odot M) \cdot \gamma,
\end{equation}
where \(\odot\) is the Hadamard product, \(L \in \mathbb{R}^{E \times E}\) is a learnable inhibition weight matrix, randomly initialized (with a standard deviation of 0.1) and set to \(0\) along the diagonal to prevent self-inhibition. \(M \in \mathbb{R}^{E \times E}\) is the inhibition mask matrix, and \(\gamma\) is the inhibition strength coefficient. This competition-inhibition mechanism ensures that when an expert neuron fires, it suppresses the activation probability of other expert neurons, thereby achieving competitive selection.

\textbf{Refractory period.} During the competitive selection process, high-frequency neurons may undergo continuous triggering. This can cause the Top-K selection to degenerate into a monopoly by a single expert, leading to expert collapse. To avoid expert collapse, we introduce a refractory period. Under this mechanism, neuronal firing follows the following threshold-triggered mechanism:
\begin{equation}
\label{equ:refractory1}
    S[t]=
    \begin{cases}
        1 \ \ \text{if} \ V[t]\ge V_{th}[t] \ \text{and} \ R[t]\le0 \\
        0 \ \ \text{otherwise}
    \end{cases},
\end{equation}
where \(S[t] \in \{0,1\}^{B \times E}\) denotes the spike firing state, while \(R[t] \in \mathbb{R}^{B \times E}\) represents the refractory period counter. After a neuron fires a spike, its membrane potential resets and enters the refractory period (meanwhile, the other neurons continue to fire):
\begin{equation}
\label{equ:refractory2}
    V[t] \leftarrow 0 \ \text{,} \ R[t] \leftarrow t_{\text{refractory}} \ \text{when} \ S[t]=1.
\end{equation}

The refractory period counter decrements over time, i.e., $R[t+1]=\text{max}(0,R[t]-1)$, ensuring neurons undergo a brief recovery period after firing a spike. To confine membrane potential and count values within reasonable bounds, we employ clamp operations to non-negate membrane potential and count values. This prevents erroneous masking resulting from negative count values.

\textbf{Expert selection and weight assignment.} To achieve expert weight allocation based on spike count, we emulate the brain's multi-timescale characteristics \citep{yang2024selection} by introducing wta-timesteps $t$ decoupled from the backbone's $T$. For an input tensor \(X \in \mathbb{R}^{T \times B \times N \times C}\), \(T\) represents the extrinsic temporal dimension of the backbone model, characterizing the temporal structure of learned and encoded inputs. Meanwhile, \(t\) represents the simulation time step within the Router, used to compute the temporal count \(c_i\) via neurodynamics, which then informs the Top-K selection. Therefore, the spike count for each expert is:
\begin{equation}
\label{equ:spike_count}
    c_i=\sum^t_{t=1}S_{b,i}[t], \ \text{for} \ i=1,2,...,E,
\end{equation}
where \(i\) denotes the number of experts, \(S_{b,i}\) represents the number of spike issued, and \(t\) is the simulation time step. Employing Top-K selection, the \(K\) experts with the highest spike counts are chosen: \(\mathcal{E}_{\text{selected}} = \text{TopK}_{i \in [E]} c_i\). Expert weights are allocated proportionally based on spike counts:
\begin{equation}
\label{equ:expert_weights}
    w_i=
    \begin{cases}
        \frac{c_i+\varepsilon}{\sum_{j \in \mathcal{E}_{selected}}(c_j+\varepsilon)} \ \ \text{if} \ i \in \mathcal{E}_{selected} \\
        0, \ \ \text{otherwise}
    \end{cases}.
\end{equation}

Here, $\varepsilon$ represents a small random perturbation used to break ties when experts compete. The k-WTA Router's algorithm flow is shown in the Appendix \ref{suppl:algorithm workflow}. The process of competitive selection by expert neurons is illustrated in the figure \ref{fig:k-WTA Router} (b) and (c). We demonstrate that the routing and expert-selection processes are spike-driven (proof in Appendix\ref{suppl:proof spike-driven}), and by evaluating inference latency, we show that the introduced wta-timesteps does not create a computational bottleneck (Appendix \ref{suppl:inference latency and throughput analysis}).

\subsection{Two-stage Missing Modality Modeling Strategy}
\begin{wrapfigure}{r}{0.55\textwidth} 
    \centering
    \vspace{-4mm}
    \includegraphics[width=\linewidth]{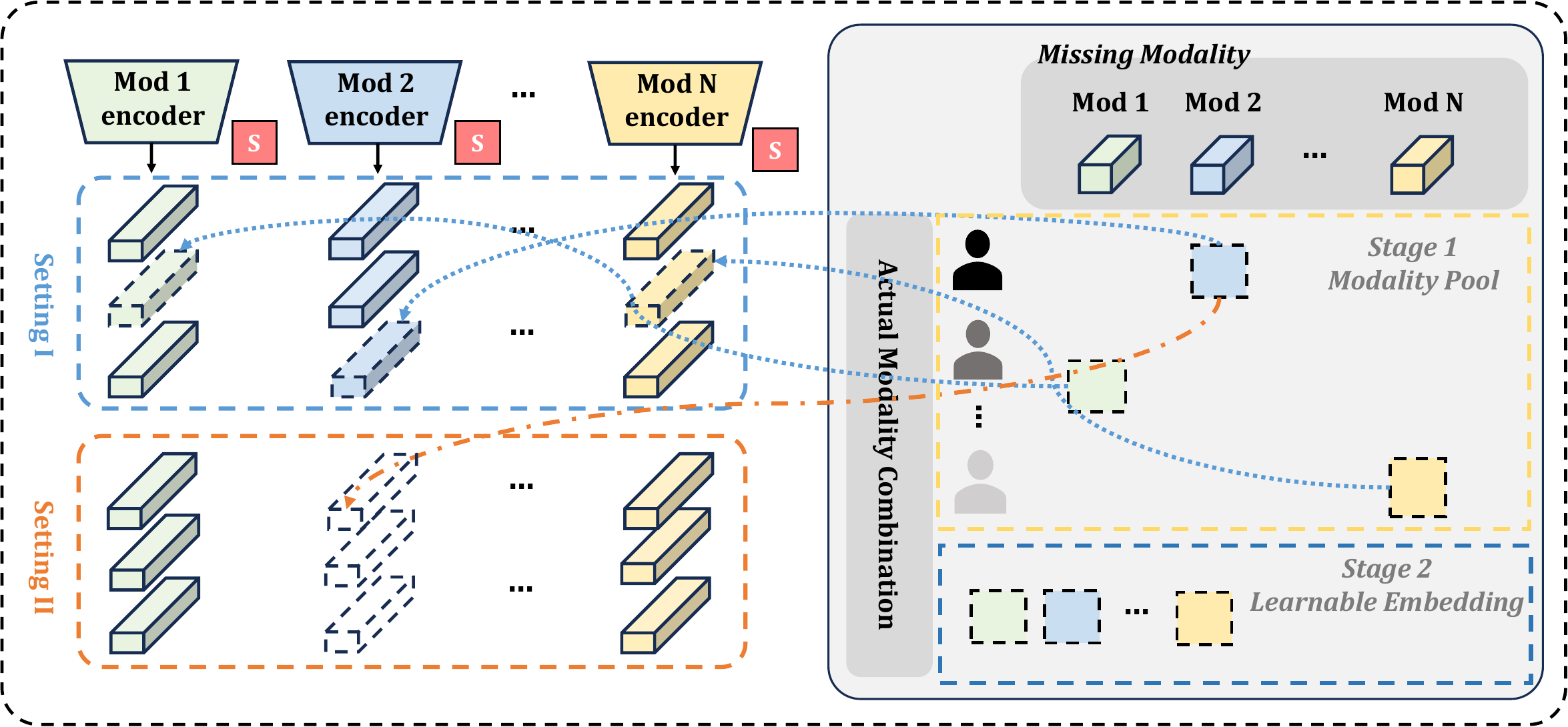}
    \vspace{-7mm}
    \caption{Two-stage missing modality modeling.}
    \vspace{-5mm}
    \label{fig:two-stage}
\end{wrapfigure}

Previous work often employs filling or interpolating for the corresponding missing modality. However, this approach becomes cumbersome when considering interactions between modalities. We consider modality combinations under spike coding to ensure robust and flexible training as well as effective handling of missing modalities. Filling these modalities must be handled in such scenarios, rather than applying globally learned representation embeddings for each modality without considering the observed context.

We propose a two-stage missing modality modeling strategy (Figure \ref{fig:two-stage}) that handles missing modalities through the adaptive fusion of a learnable modality pool and learnable embeddings. \textbf{Stage 1}: We define a binary mask \(o_{i,n}\), where \(o_{i,n}\) is defined as:
\begin{equation}
\label{equ:stage1-1}
    o_{i,n}=
    \begin{cases}
        1 &\text{if modality is missing for sample} \\
        0 &\text{otherwise}
    \end{cases},
\end{equation}
where $o_{i,n} = 1$ indicates that the modality \(M_i \in \mathbb{R}^d\) is missing, while \(o_{i,n} = 0\) indicates that the modality is present. To ensure that at least one modality is retained for each sample, we enforce a constraint:
\begin{equation}
\label{equ:stage1-2}
    \forall i \in \{1,2,...,N\}:\sum^M_{n=1}(1-o_{i,n}) \ge 1.
\end{equation}

The modality pool is defined as \(\mathcal{P}_k = \{M_{i,n}: o_{i,n}=0, \ i \in \mathcal{D}_{\text{train}} \}\), containing all existing modalities in the training set where \(o_{i,n} = 1\). When modality \(M_i\) is missing, we generate the imputed modality \(\bar{M}_i = \frac{1}{K}\sum^K_{n=1} M_{i,n}\) by sampling \(K\) instances from the modality pool \(M_{i,n} = [M_{i,0}, M_{i,1}, ..., M_{i,n}]\), where \(M_{i,n}\) represents the \(i\)-th modality in the \(n\)-th instance of the pool. 

\textbf{Stage 2}: When the modality pool cannot provide a suitable imputed modality (e.g., when the pool is empty, data quality is poor, or no matching contextual is available), a learnable embedding is used as the imputed modality. Given the number of modality combinations not fully observed, the total number of modalities in the modality pool is \(\sum^{|M|-1}_{m=1} \bigl(\begin{smallmatrix} |M| \\ m \end{smallmatrix}\bigr)=2^{|M|}-1\). Through the modality pool, the cascaded embeddings of all modalities for sample \(n\), \(h_n = [e^1_n, e^2_n, ..., e^i_n]\), are represented as follows:
\begin{equation}
\label{equ:stage2}
    e^i_n=
    \begin{cases}
        \text{Encoder}^i(n) &\text{if} \ o_{i,n}=0 \\
        M_{i,n} &\text{if} \ o_{i,n}=1
    \end{cases},
\end{equation}
where \(e^i_n \in \mathbb{R}^d\) represents the embedding of the modality \(i\) from the sample \(n\), with \(d\) being the hidden dimension. By doing so, the encoders for each modality can be trained without encountering unobserved, incomplete modalities. This approach simultaneously accounts for interactions between modalities and global representational information.

\section{Experiments}
\subsection{Experiments Settings}
\label{sec: experiments settings}

\textbf{Datasets.} Given the generality of our approach, we evaluate it on the following datasets: ImageNet-1k \citep{deng2009imagenet}, GLUE \citep{wang2018glue}, CMU-MOSI \citep{zadeh2016mosi}, CMU-MOSEI \citep{zadeh2018multimodal}, and UrbanSound8K-AV \citep{guo2023transformer}. Detailed informations are provided in Appendix \ref{suppl:datasets and task description}, while results on CMU-MOSI are available in Table \ref{tab:Experiment Setting I: CMU-MOSI} and \ref{tab:Experiment Setting II: CMU-MOSI}. We conducted experiments on CMU-MOSEI and UrbanSound-AV, with results shown in Appendix \ref{supp:results on CMU-MOSEI} and \ref{supp:results on UrbanSound8k-AV}.

\begin{wrapfigure}{r}{0.45\textwidth} 
    \centering
    \vspace{-5mm}
    \includegraphics[width=\linewidth]{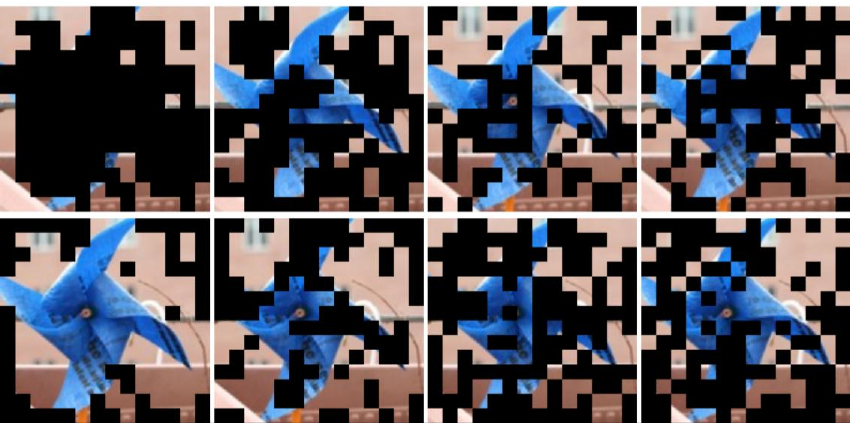}
    \vspace{-5mm}
    \caption{Visualization of routers as masks. The mask position (black) indicates router of 0 here. We show the dynamic sparsity for the selected experts (vertical) and time-steps (horizontal).}
    \vspace{-5mm}
    \label{fig:router as masks}
\end{wrapfigure}

\textbf{Experimental Settings.} On ImageNet-1K and GLUE, we evaluated its effectiveness by directly replacing the FFN layer in the baseline models with the k-WTA Router. To further assess SpikeMoE's robustness under missing modalities in real-world scenarios, we conducted extensive testing across three multimodal datasets. To simulate genuine modal deficiencies, we adopted the following two experimental settings: \textbf{\textit{Setting I: Random Modality Dropout}.} A fixed percentage of modalities are randomly discarded for each instance, while ensuring at least one modality is retained for each instance in both the training and testing sets. \textbf{\textit{Setting II: Asymmetric Modality Missingness}.} Half of the modalities are randomly discarded during training, while testing is performed only when each instance contains one or two modalities. For detailed configuration, see Appendix \ref{suppl:experiment details}

\subsection{Primary Results}
\label{sec:primary results}

\subsubsection{Effectiveness of k-WTA Router across Tasks}

\textbf{Results on ImageNet-1K.} We replaced the FFN layer in the Spiking Transformer with a k-WTA Router and compared the improved model against the baseline. As shown in Table \ref{tab:imagenet}, the k-WTA Router consistently outperforms the baseline with slightly fewer model parameters. For instance, Spikformer-8-384 with the k-WTA Router outperformed the baseline by 2.71\%, while Spike-driven Transformer-8-384 with the k-WTA Router achieved an improvement of 2.24\%. Similar findings were observed on Spikingformer. We conducted a visual analysis of the conditional computation characteristics of the k-WTA Router (4 experts, Top-k=2). As shown in the Figure~\ref{fig:router as masks}, for the same image, each expert assigns different computational regions to the routing process.

\begin{table}[htbp]
    \centering
    \vspace{-4mm}
    \caption{Results on ImageNet. Model-\(L\)-\(D\) represents a model with \(L\) encoder blocks and \(D\) channels.}
    \label{tab:imagenet}
    \resizebox{\textwidth}{!}{
    \begin{tabular}{llcccc}
        \toprule
        \textbf{Methods} & \textbf{Architecture} & \textbf{Param (M)} & \begin{tabular}{@{}c@{}}\textbf{Energy Con-} \\ \textbf{sumption (mJ)}\end{tabular} & \textbf{Time Step} & \textbf{Top-1 Acc (\%)} \\
        \midrule
        \multirow{2}{*}{Spikformer \citep{zhou2022spikformer}}
        & Spikformer-8-384 & 16.81 & 12.43 & 4 & 70.24 \\
        & Spikformer-8-512 & 29.68 & 18.82 & 4 & 73.38 \\
        \multirow{2}{*}{Spikformer+SEMM \citep{zhou2024spiking}}
        & Spikformer-8-384 & 16.05 & 11.50* & 4 & 72.86 \\
        & Spikformer-8-512 & 28.22 & 17.63* & 4 & 75.93 \\
        \rowcolor{kwtabg} & Spikformer-8-384 & \textbf{16.03} & \textbf{11.39} & 4 & \textbf{73.22} \\
        \rowcolor{kwtabg} \multirow{-2}{*}{\textbf{Spikformer+k-WTA Router}} 
        & Spikformer-8-512 & \textbf{28.19} & \textbf{16.72} & 4 & \textbf{75.97} \\
        \midrule
        \multirow{2}{*}{Spike-driven Transformer \citep{yao2023spike}}
        & SD-Transformer-8-384 & 16.81 & 3.90 & 4 & 72.28 \\
        & SD-Transformer-8-512 & 29.68 & 4.50 & 4 & 74.57 \\
        \multirow{2}{*}{SD-Transformer+SEMM \citep{zhou2024spiking}}
        & SD-Transformer-8-384 & 16.05 & 3.75* & 4 & 73.93 \\
        & SD-Transformer-8-512 & 28.22 & 4.23* & 4 & 76.62 \\
        \rowcolor{kwtabg} & SD-Transformer-8-384 & \textbf{16.03} & \textbf{3.54} & 4 & \textbf{74.52} \\
        \rowcolor{kwtabg} \multirow{-2}{*}{\textbf{SD-Transformer+k-WTA Router}} 
        & SD-Transformer-8-512 & \textbf{28.19} & \textbf{4.13} & 4 & \textbf{77.49} \\
        \midrule
        \multirow{2}{*}{Spikingformer \citep{zhou2023spikingformer}}
        & Spikingformer-8-384 & 16.81 & 4.69 & 4 & 72.45 \\
        & Spikingformer-8-512 & 29.68 & 7.46 & 4 & 74.79 \\
        \multirow{2}{*}{Spikingformer+SEMM \citep{zhou2024spiking}}
        & Spikingformer-8-384 & 16.05 & 4.47* & 4 & 73.58 \\
        & Spikingformer-8-512 & 28.22 & 6.94* & 4 & 76.03 \\
        \rowcolor{kwtabg} & Spikingformer-8-384 & \textbf{16.03} & \textbf{4.27} & 4 & \textbf{74.37} \\
        \rowcolor{kwtabg} \multirow{-2}{*}{\textbf{Spikingformer+k-WTA Router}}  
        & Spikingformer-8-512 & \textbf{28.19} & \textbf{6.63} & 4 & \textbf{76.92} \\
        \bottomrule
    \end{tabular}}
    \parbox{\textwidth}{\footnotesize \textit{Note:} * denotes our reproduced results.}
\vspace{-4mm}
\end{table}

\textbf{Results on GLUE Benchmark.} For comparison purposes on the GLUE Benchmark, we replace the FFN layer in SpikeLM with k-WTA Router and compare the enhanced model against baselines. The ANN baselines include BERT \citep{devlin2019bert} and Q2BERT (with 2-bit weights and 8-bit activations) \citep{zhang2020ternarybert}, while the SNN baselines consist of SpikeBERT \citep{lv2023spikebert} and directly trained Spikingformer \citep{zhou2023spikingformer}. Following \citep{xing2024spikelm}, we also implement spike-driven BERTs using PSN \citep{fang2023parallel} and LIF \citep{gerstner2014neuronal} neuron. As shown in Table \ref{tab:glue_results}, the k-WTA Router improves SpikeLM performance by 1.2\%. Furthermore, it narrows the performance gap with BERT\textsubscript{base} to 5.5\%, significantly improving upon the original gaps of 6.7\% for SpikeLM and 28.3\% for LIF-BERT.

\begin{table}[htbp]
    \centering
    \vspace{-4mm}
    \caption{Performance comparison on GLUE benchmark.}
    \label{tab:glue_results}
    \resizebox{\textwidth}{!}{
        \begin{tabular}{lccccccccccc}
            \toprule
            \textbf{Model} & \textbf{Energy (mJ)} & \textbf{Time} & \textbf{MNLI\textsubscript{-m/mm}} & \textbf{QQP\textsubscript{F1}} & \textbf{QNLI} & \textbf{SST-2} & \textbf{CoLA} & \textbf{STS-B} & \textbf{MRPC\textsubscript{F1}} & \textbf{RTE} & \textbf{Avg.} \\
            \midrule
            BERT\textsubscript{base} \citep{devlin2019bert}     & 51.41 & / & 83.8/83.4 & 90.5 & 90.7 & 92.3 & 60.0 & 89.4 & 89.8 & 69.3 & 83.2 \\
            BERT\textsubscript{3L} \citep{devlin2019bert}     & 12.90 & / & 77.1/77.1 & 85.2 & 85.8 & 88.1 & 31.7 & 85.7 & 86.4 & 66.4 & 75.9 \\
            Q2BERT \citep{zhang2020ternarybert}       & /     & / & 47.2/47.3 & 67.0 & 61.3 & 80.6 & 0.0  & 4.7  & 81.2 & 52.7 & 49.1 \\
            ELMo  \citep{peters-etal-2018-deep}        & /     & / & 68.6/-    & 86.2 & 71.1 & 91.5 & 44.1 & 70.4 & 76.6 & 53.4 & 70.2 \\
            SpikeBERT \citep{lv2023spikebert}    & 14.30 & 4 & 71.4/71.0 & 68.2 & 66.4 & 85.4 & 16.9 & 18.7 & 82.0 & 57.5 & 59.7 \\
            LIF-BERT* \citep{gerstner2014neuronal, xing2024spikelm} & /     & 4 & 35.4/35.2 & 0.0  & 50.5 & 50.9 & 0.0  & 0.0  & 81.2 & 52.7 & 34.6 \\
            PSN-BERT* \citep{fang2023parallel, xing2024spikelm} & /     & 4 & 35.4/35.2 & 0.0  & 50.5 & 50.9 & 0.0  & 6.8  & 81.2 & 52.7 & 34.7 \\
            LIF-BERT \citep{gerstner2014neuronal, xing2024spikelm}      & 7.98  & 4 & 56.8/55.2 & 70.0 & 60.6 & 80.6 & 14.6 & 20.0 & 82.3 & 53.8 & 54.9 \\
            Spikingformer \citep{zhou2023spikingformer} & /     & 1 & 67.8/68.6 & 79.3 & 74.6 & 82.7 & 16.7 & 72.3 & 83.0 & 58.8 & 67.1 \\
            Spikingformer \citep{zhou2023spikingformer} & /     & 4 & 70.2/70.6 & 80.9 & 79.5 & 83.9 & 12.8 & 77.0 & 83.0 & 62.1 & 68.9 \\
            SpikeLM \citep{xing2024spikelm}      & 13.74 & 4 & 77.1/77.2 & 83.9 & 85.3 & 87.0 & 38.8 & 84.9 & 85.7 & 69.0 & 76.5 \\
            \rowcolor{kwtabg} \textbf{SpikeLM+k-WTA Router} & \textbf{12.45} & 4 & \textbf{77.2/77.4} & \textbf{85.6} & \textbf{86.6} & \textbf{88.2} & \textbf{42.2} & \textbf{86.4} & \textbf{86.2} & \textbf{69.7} & \textbf{77.7} \\
            \bottomrule
        \end{tabular}
        }
\end{table}

\subsubsection{Robustness of SpikeMoE to Missing Modalities}
\begin{wrapfigure}{r}{0.4\textwidth}
\centering 
\includegraphics[width=\linewidth]{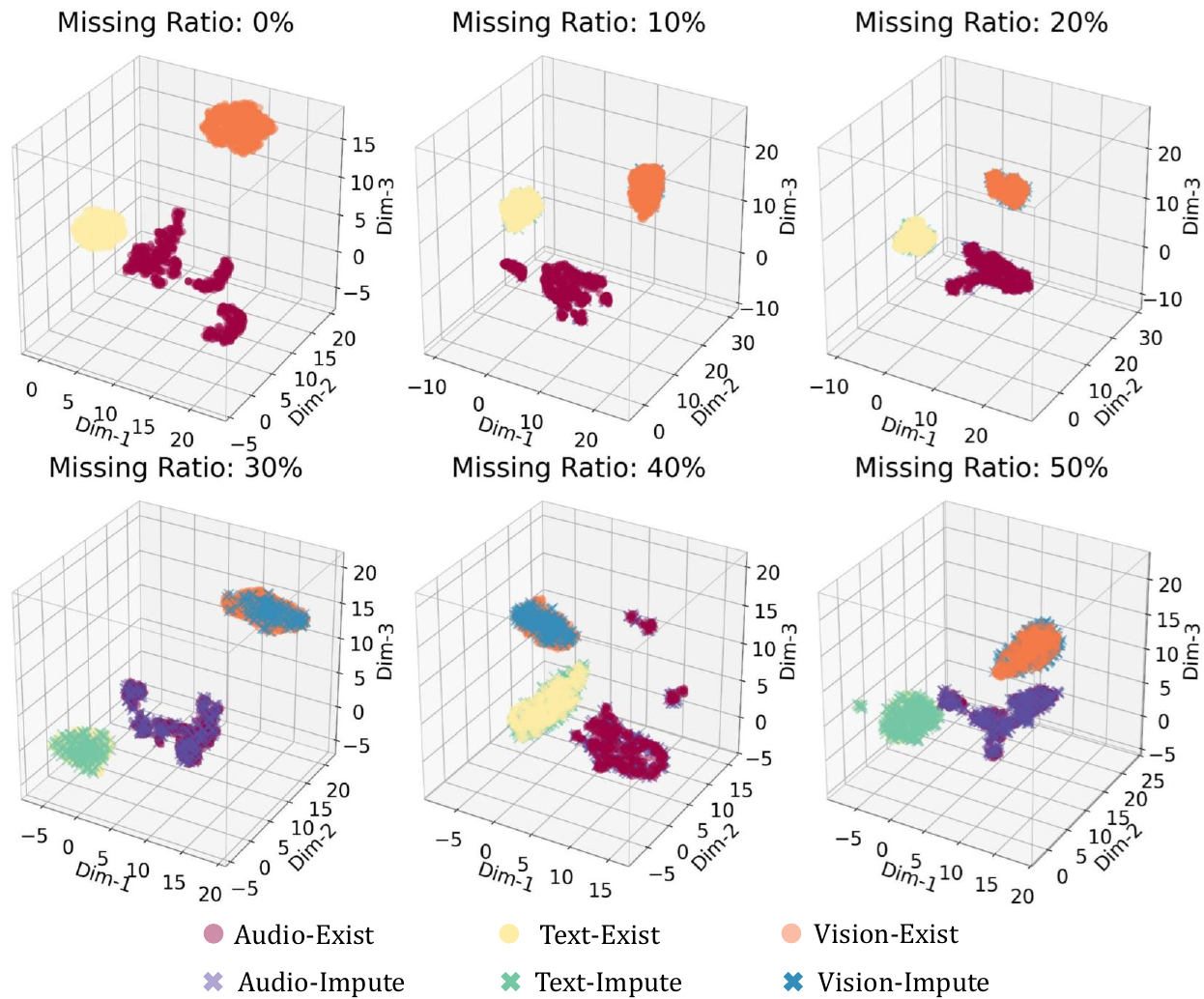}
\vspace{-6mm}
\caption{UMAP visualization of multi-modal embedding space: CMU-MOSI.}
\vspace{-5mm}
\label{fig:umap}
\end{wrapfigure}

\textbf{Experiment Setting I:} In Table~\ref{tab:Experiment Setting I: CMU-MOSI} and Table~\ref{tab:Experiment Setting I: CMU-MOSEI}, we implement Random Modality Dropout to study the sensitivity of model to missing modality. In this setting, we empirically observe that methods employing missing modal interpolation typically exhibit greater robustness under high levels of missing data, such as FuseMoE \citep{han2024fusemoe} and our proposed approach. Although ShaSpec \citep{wang2023multi} also estimates missing modes by sharing information, it is weak on the backbone model and difficult to optimize with multi-task objectives. On the other hand, compared to ANNs methods, our proposed approach demonstrates competitive performance across all metrics across most settings of CMU-MOSEI and CMU-MOSI; among SNNs methods, our approach achieves state-of-the-art performance. We present the UMAP projection of multimodal embeddings on the two datasets in Fig. \ref{fig:umap} and \ref{fig:umap_mosei}, demonstrating that the impute modalities follow the same distribution as the original modalities. This validates the effectiveness of our method.
\textbf{Experiment Setting II:} In Tables ~\ref{tab:Experiment Setting II: CMU-MOSI} and ~\ref{tab:Experiment Setting II: CMU-MOSEI}, we investigate which modality combinations are significant and how the resilience of our proposed method varies across different distributions under missing modalities. We find that in our proposed method, single modalities exhibit similar performance on F1 score and AUC during the inference stage, indicating the effectiveness of the missing modality filling strategy. Specifically, our proposed method also demonstrates robustness to missing modalities across different combinations, meaning it maintains competitive performance regardless of which modality is absent. Furthermore, when evaluating the performance of all methods collectively, the video-audio modality combination yields poorer results, potentially due to weaker interactivity between these two modalities.

\begin{table*}[t]
\centering
\setlength{\tabcolsep}{1.4pt}
\renewcommand{\arraystretch}{1.1}
\setlength{\fboxsep}{1.0pt}
\caption{Experiment Setting I: Main results on CMU-MOSI. Top-3 results are highlighted as \smash{\colorbox{firstcolor}{first}}, \smash{\colorbox{secondcolor}{second}} and \smash{\colorbox{thirdcolor}{third}}.}
\label{tab:Experiment Setting I: CMU-MOSI}
\resizebox{\textwidth}{!}{%
\begin{tabular}{cccccccccc}
\toprule
\multirow{3}{*}{\shortstack[c]{\textbf{Missing}\\\textbf{Ratio}}} &
\multirow{3}{*}{\textbf{Metric}} &
\multicolumn{4}{c}{\textbf{ANNs}} &
\multicolumn{3}{c}{\textbf{SNNs}} &
\multicolumn{1}{c}{\textbf{Ours}} \\
\cmidrule(lr){3-6}\cmidrule(lr){7-9} \cmidrule(lr){10-10}
& &
\shortstack[c]{\textbf{ShaSpec}\\\citep{wang2023multi}} &
\shortstack[c]{\textbf{TF}\\\citep{zadeh2017tensor}} &
\shortstack[c]{\textbf{mmFormer}\\\citep{zhang2022mmformer}} &
\shortstack[c]{\textbf{FuseMoE}\\\citep{han2024fusemoe}} &
\shortstack[c]{\textbf{Weight Attention}\\\citep{liu2022event}} &
\shortstack[c]{\textbf{SCA}\\\citep{guo2023transformer}} &
\shortstack[c]{\textbf{S-CMRL}\\\citep{he2025enhancing}} &
\shortstack[c]{\textbf{SpikeMoE}} \\
\midrule
\multirow{2}{*}{0\%}
 & F1  & 41.78$\pm$0.64 & 43.28$\pm$0.36 & 50.22$\pm$0.20 & \smash{\colorbox{secondcolor}{57.02$\pm$0.39}} & \smash{\colorbox{thirdcolor}{50.41$\pm$0.44}} & 46.95$\pm$0.21 & 49.01$\pm$0.29 & \smash{\colorbox{firstcolor}{59.59$\pm$0.23}} \\
 & AUC & 65.08$\pm$0.39 & 64.48$\pm$0.51 & \smash{\colorbox{thirdcolor}{74.62$\pm$0.49}} & \smash{\colorbox{secondcolor}{77.26$\pm$0.20}} & 72.09$\pm$0.38 & 70.33$\pm$0.46 & 73.94$\pm$0.50 & \smash{\colorbox{firstcolor}{78.48$\pm$0.34}} \\
\midrule
\multirow{2}{*}{10\%}
 & F1  & 43.40$\pm$0.74 & 40.14$\pm$0.13 & \smash{\colorbox{thirdcolor}{48.66$\pm$1.20}} & \smash{\colorbox{secondcolor}{55.81$\pm$0.45}} & 48.19$\pm$0.52 & 46.81$\pm$0.71 & 47.50$\pm$0.06 & \smash{\colorbox{firstcolor}{56.51$\pm$0.12}} \\
 & AUC & 65.45$\pm$0.56 & 61.54$\pm$0.66 & 73.21$\pm$0.76 & \smash{\colorbox{secondcolor}{77.05$\pm$0.93}} & \smash{\colorbox{thirdcolor}{74.68$\pm$0.17}} & 70.51$\pm$0.02 & 72.12$\pm$0.27 & \smash{\colorbox{firstcolor}{77.45$\pm$0.67}} \\
\midrule
\multirow{2}{*}{20\%}
 & F1  & 39.15$\pm$0.13 & 42.64$\pm$0.30 & \smash{\colorbox{thirdcolor}{48.51$\pm$1.15}} & \smash{\colorbox{secondcolor}{53.42$\pm$0.58}} & 47.46$\pm$0.02 & 47.43$\pm$0.29 & 47.35$\pm$0.35 & \smash{\colorbox{firstcolor}{53.73$\pm$0.92}} \\
 & AUC & 60.87$\pm$0.29 & 65.52$\pm$0.98 & \smash{\colorbox{firstcolor}{74.16$\pm$0.71}} & 73.05$\pm$0.36 & \smash{\colorbox{thirdcolor}{73.54$\pm$0.81}} & 68.22$\pm$0.34 & 72.31$\pm$0.74 & \smash{\colorbox{secondcolor}{74.02$\pm$0.44}} \\
\midrule
\multirow{2}{*}{30\%}
 & F1  & 39.54$\pm$0.17 & 40.42$\pm$0.75 & 46.66$\pm$1.24 & \smash{\colorbox{firstcolor}{51.78$\pm$0.51}} & \smash{\colorbox{thirdcolor}{46.87$\pm$0.03}} & 44.61$\pm$0.08 & 44.01$\pm$1.01 & \smash{\colorbox{secondcolor}{51.04$\pm$0.02}} \\
 & AUC & 60.80$\pm$0.44 & 60.57$\pm$0.99 & 71.25$\pm$0.57 & \smash{\colorbox{secondcolor}{72.38$\pm$0.56}} & \smash{\colorbox{thirdcolor}{72.31$\pm$0.11}} & 68.48$\pm$1.03 & 70.21$\pm$0.05 & \smash{\colorbox{firstcolor}{73.15$\pm$0.35}} \\
\midrule
\multirow{2}{*}{40\%}
 & F1  & 35.32$\pm$0.32 & 38.69$\pm$0.42 & 42.22$\pm$0.44 & \smash{\colorbox{secondcolor}{48.78$\pm$0.42}} & \smash{\colorbox{thirdcolor}{48.45$\pm$0.20}} & 41.81$\pm$0.37 & 42.11$\pm$0.64 & \smash{\colorbox{firstcolor}{48.99$\pm$0.51}} \\
 & AUC & 59.58$\pm$0.13 & 61.62$\pm$0.07 & 68.44$\pm$0.14 & 68.61$\pm$0.63 & \smash{\colorbox{secondcolor}{68.93$\pm$0.36}} & 66.52$\pm$0.12 & \smash{\colorbox{thirdcolor}{68.74$\pm$0.39}} & \smash{\colorbox{firstcolor}{69.02$\pm$0.32}} \\
\midrule
\multirow{2}{*}{50\%}
 & F1  & 36.39$\pm$0.97 & 38.46$\pm$1.09 & 41.14$\pm$0.31 & \smash{\colorbox{thirdcolor}{42.55$\pm$0.34}} & 41.65$\pm$0.31 & \smash{\colorbox{secondcolor}{43.05$\pm$0.65}} & 42.03$\pm$0.68 & \smash{\colorbox{firstcolor}{43.54$\pm$0.03}} \\
 & AUC & 55.60$\pm$0.12 & 60.49$\pm$0.80 & 61.87$\pm$0.67 & \smash{\colorbox{secondcolor}{67.28$\pm$0.28}} & \smash{\colorbox{thirdcolor}{65.81$\pm$0.05}} & 64.49$\pm$0.43 & 65.71$\pm$0.43 & \smash{\colorbox{firstcolor}{67.30$\pm$0.71}} \\
\bottomrule
\end{tabular}%
}
\end{table*}

\begin{table*}[t]
\centering
\setlength{\tabcolsep}{1.4pt}
\renewcommand{\arraystretch}{1.1}
\setlength{\fboxsep}{1.0pt}
\vspace{-4mm}
\caption{Experiment Setting II: Main results on CMU-MOSI. Top-3 results are highlighted as \smash{\colorbox{firstcolor}{first}}, \smash{\colorbox{secondcolor}{second}} and \smash{\colorbox{thirdcolor}{third}}.}
\label{tab:Experiment Setting II: CMU-MOSI}
\resizebox{\textwidth}{!}{%
\begin{tabular}{cccccccccccc}
\toprule
\multicolumn{3}{c}{\textbf{Testing Modality}} &
\multirow{3}{*}{\textbf{Metric}} &
\multicolumn{4}{c}{\textbf{ANNs}} &
\multicolumn{3}{c}{\textbf{SNNs}} &
\multicolumn{1}{c}{\textbf{Ours}} \\
\cmidrule(lr){1-3}\cmidrule(lr){5-8}\cmidrule(lr){9-11}\cmidrule{12-12}
\textbf{Video} & \textbf{Text} & \textbf{Audio} & &
\shortstack[c]{\textbf{ShaSpec}\\\citep{wang2023multi}} &
\shortstack[c]{\textbf{TF}\\\citep{zadeh2017tensor}} &
\shortstack[c]{\textbf{mmFormer}\\\citep{zhang2022mmformer}} &
\shortstack[c]{\textbf{FuseMoE}\\\citep{han2024fusemoe}} &
\shortstack[c]{\textbf{Weight Attention}\\\citep{liu2022event}} &
\shortstack[c]{\textbf{SCA}\\\citep{guo2023transformer}} &
\shortstack[c]{\textbf{S-CMRL}\\\citep{he2025enhancing}} &
\textbf{SpikeMoE} \\
\midrule
\checkmark & \checkmark & & F1 &
41.27$\pm$0.81 & 34.96$\pm$0.72 &
\smash{\colorbox{thirdcolor}{48.25$\pm$0.11}} &
30.51$\pm$2.53 & 25.75$\pm$0.53 & 47.55$\pm$0.56 &
\smash{\colorbox{secondcolor}{48.86$\pm$0.34}} &
\smash{\colorbox{firstcolor}{69.56$\pm$0.23}} \\
& & & AUC &
61.28$\pm$0.88 & 59.37$\pm$1.12 &
\smash{\colorbox{thirdcolor}{72.72$\pm$0.08}} &
55.35$\pm$1.11 & 54.39$\pm$0.12 & 71.98$\pm$0.43 &
\smash{\colorbox{secondcolor}{73.22$\pm$0.06}} &
\smash{\colorbox{firstcolor}{74.29$\pm$0.18}} \\
\midrule
\checkmark & & \checkmark & F1 &
\smash{\colorbox{thirdcolor}{36.99$\pm$2.03}} &
34.83$\pm$0.54 & 35.79$\pm$0.55 & 24.29$\pm$4.23 &
22.18$\pm$0.37 &
\smash{\colorbox{secondcolor}{46.97$\pm$0.08}} &
23.82$\pm$0.39 &
\smash{\colorbox{firstcolor}{47.81$\pm$0.61}} \\
& & & AUC &
55.79$\pm$2.53 &
\smash{\colorbox{thirdcolor}{57.87$\pm$0.32}} &
52.21$\pm$0.31 & 56.24$\pm$1.96 & 50.45$\pm$0.08 &
\smash{\colorbox{secondcolor}{68.56$\pm$0.23}} &
51.55$\pm$0.28 &
\smash{\colorbox{firstcolor}{69.04$\pm$0.18}} \\
\midrule
& \checkmark & \checkmark & F1 &
43.72$\pm$1.64 & 37.25$\pm$0.53 & 45.53$\pm$0.33 &
20.68$\pm$2.14 &
\smash{\colorbox{secondcolor}{49.40$\pm$0.23}} &
19.18$\pm$0.47 &
\smash{\colorbox{thirdcolor}{48.09$\pm$0.21}} &
\smash{\colorbox{firstcolor}{49.47$\pm$0.38}} \\
& & & AUC &
65.22$\pm$1.01 & 57.89$\pm$0.65 &
\smash{\colorbox{thirdcolor}{71.96$\pm$0.47}} &
54.36$\pm$1.67 &
\smash{\colorbox{secondcolor}{72.08$\pm$0.07}} &
49.40$\pm$0.28 & 68.95$\pm$0.43 &
\smash{\colorbox{firstcolor}{72.98$\pm$0.52}} \\
\midrule
\checkmark & & & F1 &
42.98$\pm$0.21 & 38.11$\pm$0.35 &
\smash{\colorbox{secondcolor}{46.49$\pm$0.91}} &
25.67$\pm$2.82 & 23.85$\pm$0.34 & 45.32$\pm$0.47 &
\smash{\colorbox{thirdcolor}{45.76$\pm$0.50}} &
\smash{\colorbox{firstcolor}{49.09$\pm$0.04}} \\
& & & AUC &
55.10$\pm$1.59 & 58.87$\pm$0.44 &
\smash{\colorbox{secondcolor}{71.01$\pm$0.53}} &
55.94$\pm$1.51 & 52.49$\pm$0.38 & 70.21$\pm$0.55 &
\smash{\colorbox{thirdcolor}{70.32$\pm$0.03}} &
\smash{\colorbox{firstcolor}{72.29$\pm$0.09}} \\
\midrule
& \checkmark & & F1 &
41.36$\pm$2.27 & 37.99$\pm$0.40 &
\smash{\colorbox{secondcolor}{47.13$\pm$0.54}} &
25.70$\pm$0.32 & 24.72$\pm$0.49 & 43.55$\pm$0.34 &
\smash{\colorbox{thirdcolor}{43.82$\pm$0.32}} &
\smash{\colorbox{firstcolor}{47.68$\pm$0.80}} \\
& & & AUC &
61.60$\pm$4.80 & 60.28$\pm$1.42 &
\smash{\colorbox{firstcolor}{73.77$\pm$0.44}} &
55.45$\pm$0.95 & 53.87$\pm$0.33 & 71.02$\pm$0.27 &
\smash{\colorbox{thirdcolor}{71.52$\pm$0.04}} &
\smash{\colorbox{secondcolor}{72.30$\pm$0.26}} \\
\midrule
& & \checkmark & F1 &
\smash{\colorbox{thirdcolor}{45.45$\pm$2.73}} &
38.79$\pm$0.56 &
\smash{\colorbox{secondcolor}{46.02$\pm$1.21}} &
25.23$\pm$4.59 & 43.52$\pm$0.26 & 42.97$\pm$0.28 &
43.79$\pm$0.12 &
\smash{\colorbox{firstcolor}{46.47$\pm$0.13}} \\
& & & AUC &
66.85$\pm$3.01 & 57.19$\pm$0.72 &
\smash{\colorbox{thirdcolor}{68.04$\pm$0.94}} &
56.04$\pm$1.91 & 54.08$\pm$0.63 & 67.06$\pm$0.02 &
\smash{\colorbox{secondcolor}{68.22$\pm$0.15}} &
\smash{\colorbox{firstcolor}{69.03$\pm$0.08}} \\
\bottomrule
\end{tabular}%
}
\vspace{-4mm}
\end{table*}

\subsection{Ablation Study}
\begin{wrapfigure}{r}{0.4\textwidth}
\vspace{-4mm}
\centering 
\includegraphics[width=\linewidth]{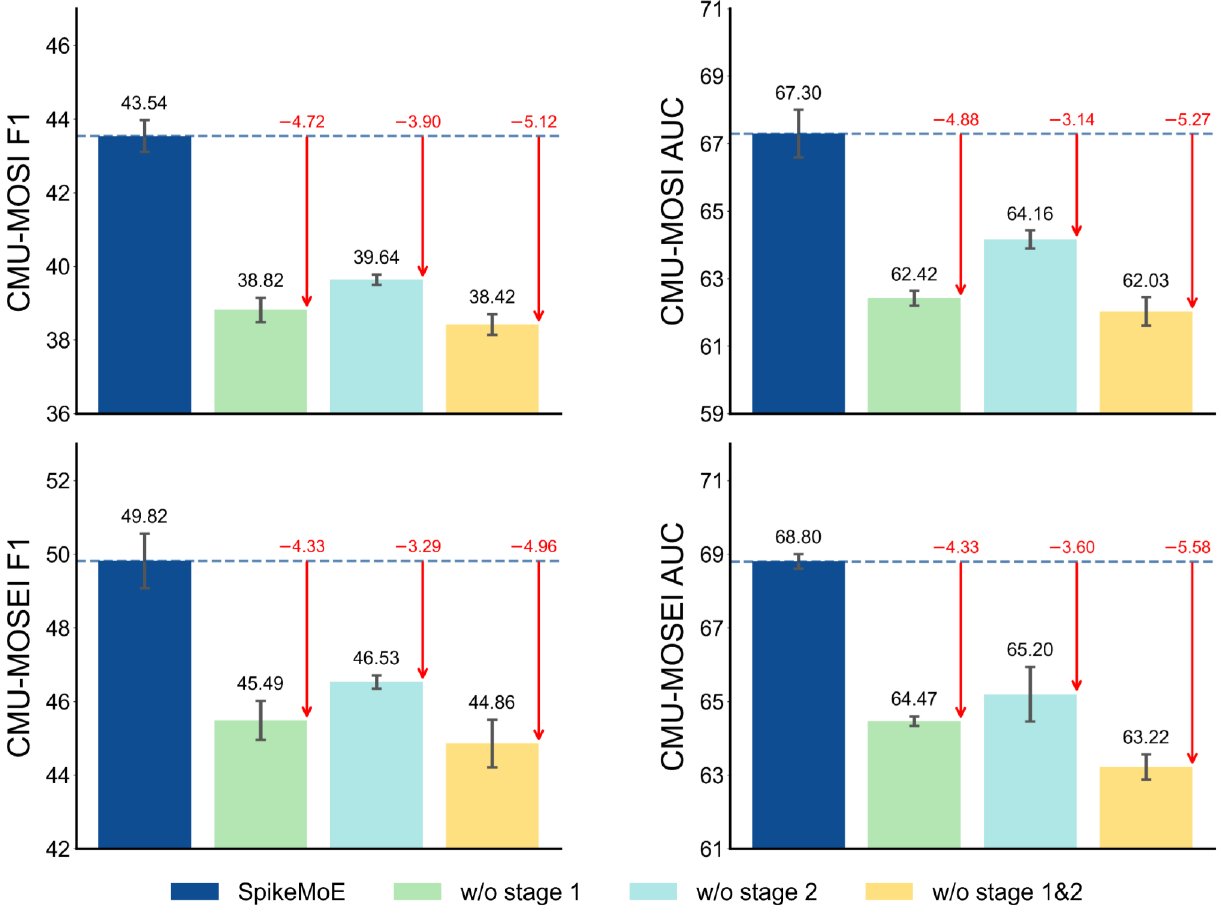}
\vspace{-7mm}
\caption{Ablation on two-stage strategy.}
\vspace{-5mm}
\label{fig:ablation}
\end{wrapfigure}

\begin{wraptable}{r}{0.5\textwidth}
  \vspace{-2em} 
  \centering
  \caption{Ablation on k-WTA Router.}
  \label{tab:ablation_kwta}
  \resizebox{\linewidth}{!}{
  \begin{tabular}{lcc}
    \toprule
    \multirow{2}{*}{\textbf{Methods}} & \multicolumn{2}{c}{\textbf{Missing Ratio=50\%}} \\
    \cmidrule(lr){2-3}
    & \textbf{F1} & \textbf{AUC} \\
    \midrule
    Fully k-WTA Router & \(43.54 \pm 0.03\) & \(67.30 \pm 0.71\) \\
    w/o Lateral Inhibition & \(41.37 \pm 0.71\) & \(66.02 \pm 0.49\) \\
    w/o Refractory Period & \(42.50 \pm 0.34\) & \(66.12 \pm 0.56\) \\
    w/o Inhibition Strength & \(40.41 \pm 0.43\) & \(64.59 \pm 0.80\) \\
    \bottomrule
  \end{tabular}}
  \vspace{-1em} 
\end{wraptable}
\textbf{Ablation on two-stage strategy.} By progressively removing components designed in our proposed method, we investigated which parts of the model are actually useful (50\% missing ratio on CMU-MOSI, as shown in Figure \ref{fig:ablation}). We sequentially discarded the two stages of the missing modality modeling strategy and replaced missing modalities with zeros. When both stages were removed, F1 and AUC decreased by 5.12\% and 5.27\% on CMU-MOSI, and by 4.96\% and 5.58\% on CMU-MOSEI. Furthermore, we observed that removing stage 1 resulted in a greater decline than stage 2, suggesting that the modal interactions captured in the modality pool may play a more critical role in missing modalality modeling.

\begin{figure*}[htbp]
    \centering
    \vspace{-2mm}
    \includegraphics[width=1\textwidth]{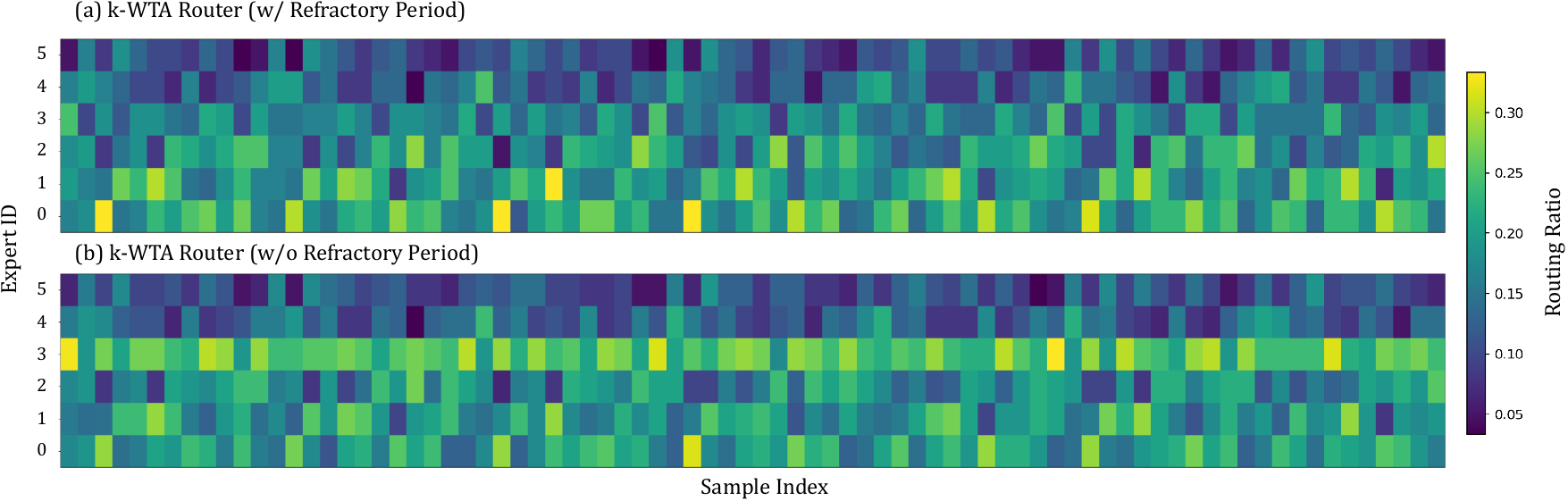}  
    \vspace{-7mm}
    \caption{Visualization of expert activation.}
    \vspace{-5mm}
    \label{fig:visualization of expert activation}
\end{figure*}

\textbf{Ablation on k-WTA Router.} To validate the internal mechanisms of the proposed k-WTA Router, we conducted ablation studies at a 50\% missing ratio on CMU-MOSI. As shown in Table \ref{tab:ablation_kwta}, discarding key components—such as lateral inhibition or the refractory period—results in consistent performance degradation. Notably, eliminating the inhibition mechanism entirely (w/o Inhibition Strength) causes the most severe drop, with F1 and AUC falling by 3.13\% and 2.71\%, respectively. Furthermore, Fig.\ref{fig:visualization of expert activation} provides an intuitive explanation for this decline: without the refractory period, the routing process suffers from "expert collapse," where a single expert (expert 3) monopolizes the activation. These results collectively confirm that our brain-inspired competition-inhibition mechanisms are essential for preserving dynamic load balance and ensuring robust representation.

\subsection{Sensitivity Analysis}

\begin{figure*}[htbp]
    \centering
    \includegraphics[width=1\textwidth]{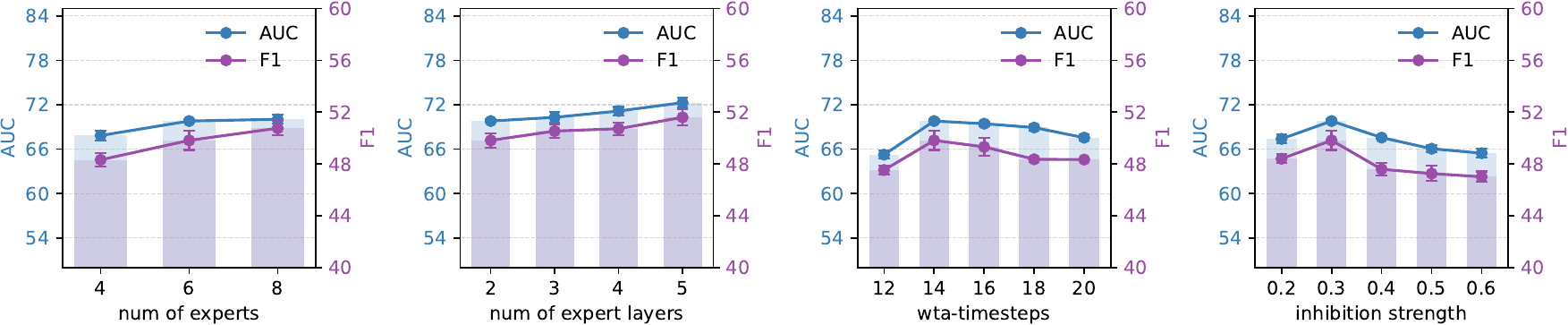}  
    \vspace{-6mm}
    \caption{Sensitivity analysis on CMU-MOSEI (Missing ratio=50\%).}
    \vspace{-5mm}
    \label{fig:sensitivity}
\end{figure*}

To demonstrate the robustness of SpikeMoE, we tested various settings of the number of experts, wta-timesteps, and inhibition strength under the configuration of missing ratio = 50\% on CMU-MOSEI. We performed the sensitivity analysis by varying the values of the corresponding hyperparameters while keeping the other hyperparameters fixed at the values used in the experiments. Figure \ref{fig:sensitivity} presents the results of different parameters, indicating that the overall metrics remain relatively stable (dissused in \ref{suppl: discussion of sensitivity analysis}). In addition, we found that performance improved as the number and depth of experts increased, indicating that SpikeMoE demonstrates promising scalability.

\subsection{Complexity Analysis}

We compared the complexity of our proposed method and baseline models in terms of OPs (SOPs in SNN and FLOPs in ANN) parameters and energy consumption on CMU-MOSI (Missing ratio=50\%). ShaSpec is the simplest model, yet its task performance is not comparable. In contrast, mmformer and S-CMRL are highly complex in terms of OPs and parameters, but their performance is not optimal. FuseMoE demonstrates performance comparable to our proposed method, yet it requires significantly more OPs and parameters. SpikeMoE is the optimal model that best balances complexity and performance. Detailed data are provided in Table \ref{tab:complexity}.


\section{Conclusion}



In this work, we propose SpikeMoE, a unified framework that integrates Spiking Neural Networks (SNNs) with the Mixture-of-Experts (MoE) architecture. To address the inherent conflict between traditional softmax gating and SNN dynamics, we introduce a brain-inspired spiking gating network based on a competition-refractory-inhibition mechanism, which achieves dynamic expert selection through sparse spiking activation. Furthermore, to enhance the model's adaptability in complex scenarios, we propose a two-stage missing-modality modeling strategy. Extensive evaluations across vision, language, and multimodal datasets demonstrate SpikeMoE's robust performance. Future work will focus on applying SpikeMoE to a broader range of tasks, scaling the framework to larger architectures, and exploring its deployment on neuromorphic hardware.

\subsection*{AI use statement}

We used generative AI tools to improve the grammar, clarity, and wording of the manuscript, including assistance in drafting these statements. We did not use these tools to develop the research methodology, implement the model, generate or process experimental data, or analyze and interpret results. The authors reviewed and revised the AI-assisted text, verified its technical content against the research record, and take full responsibility for the final manuscript.




\subsection*{Ethics statement}

This study uses established image, language, and audiovisual benchmark datasets and involved no new recruitment of human participants or collection of personal data. The CMU-MOSI and CMU-MOSEI benchmarks contain recordings of people; their use follow the dataset providers’ terms and applicable privacy requirements. Dataset composition and sentiment annotations may introduce biases that affect performance across populations.



\subsection*{Reproducibility statement}

We provide the model equations and implementation details needed to reproduce SpikeMoE in Section \ref{methods}, including the k-WTA Router’s membrane dynamics, lateral inhibition, refractory mechanism, spike-count-based expert selection, and two-stage missing-modality modeling strategy. Appendix \ref{suppl:algorithm workflow} presents pseudocode for the router, while Appendix \ref{suppl:main theorems and proofs} provides the argument for its spike-driven characteristics. Section \ref{sec: experiments settings} defines the evaluation tasks and two missing-modality settings, including how modalities are removed during training and testing. Appendix \ref{suppl:experiment details} documents the dataset partitions, evaluation metrics, modality-specific encoders, baseline methods, training objective, hardware, and energy-estimation procedure. Table \ref{tab: hyperparameter setting} reports the principal hyperparameters and random seeds. We additionally report ablations and sensitivity analyses, with further results for CMU-MOSEI and UrbanSound8K-AV in Appendices \ref{supp:results on CMU-MOSEI} and \ref{supp:results on UrbanSound8k-AV}. Together, these descriptions specify the experimental protocol and clarify that the reported energy consumption is a theoretical estimate rather than a hardware measurement.





\bibliography{iclr2027_conference}
\bibliographystyle{iclr2027_conference}

\newpage

\appendix
\section{Appendix}

\subsection{Main Propositions and Proofs}
\label{suppl:main theorems and proofs}

\subsubsection{Proof of Spike-driven Characteristics}
\label{suppl:proof spike-driven}

We prove the spike-driven characteristics of the routing and expert-selection processes in k-WTA Router based on the definition proposed by \citep{shi2024spikingresformer}.

\begin{definition}
\label{def: spike-driven}
A spiking neural network is spike-driven if the input currents of all neurons satisfy the following form:
\begin{equation}
    I_i[t]=\sum_jw_{i,j}s_j[t]=\sum_{j,s_j[t]\neq0}w_{i,j},
\end{equation}
where \(I_i[t]\) is the input current of the \(i\)-th postsynaptic neuron at time step \(t\), \(s_j[t] \in \{0, 1\}\) is the spike output of the \(j\)-th presynaptic neuron, \(w_{i,j}\) is the weight of the synaptic connection from neuron \(j\) to neuron \(i\). 
\end{definition}

\begin{proof}
The membrane potential dynamics of the k-WTA Router are described by the discrete-time equation \ref{equ:lif}. Here, \( V[t] \in \mathbb{R}^{B \times E} \) represents the membrane potential tensor, where \( B \) is the batch size and \( E \) is the number of experts. \( \tau \) denotes the membrane potential time constant, and \( I[t] \) is the input current tensor.

In the k-WTA Router, the input current \( I[t] \) originates from the spike sequence of the backbone network. Let the output of the backbone network be the input tensor \( X \in \mathbb{R}^{T \times B \times N \times C} \), where \( T \) is the backbone time step, \( B \) is the batch size, \( N \) is the sequence length or feature dimension, and \( C \) is the number of channels. In SpikeMoE, \( X \) is encoded as a binary spike sequence, with each element \( s_j[t] \in \{0,1\} \) representing the spike output of the presynaptic neuron, indexed by \( j \).

For each neuron \( i \) in the k-WTA Router (\( i=1,2,\ldots,E \)), the input current \( I_i[t] \) is computed as the weighted sum of the spike outputs from the presynaptic neurons:
\begin{equation}
    I_i[t] = \sum_j w_{i,j} s_j[t],
\end{equation} 
where \( w_{i,j} \) is the synaptic weight from presynaptic neuron \( j \) to k-WTA Router neuron \( i \), satisfying Definition 1.

Moreover, the lateral inhibition and refractory period mechanisms introduced in the k-WTA Router do not affect the spike-driven nature of the input current: 

\begin{itemize}
    \item As shown in equation \ref{equ:inhibit}, the lateral inhibition mechanism acts on the membrane potential. Here, \( S[t] \in \{0,1\}^{B \times E} \) represents the spike output of the k-WTA Router neurons, \( L \in \mathbb{R}^{E \times E} \) is the learnable inhibition weight matrix, \( M \in \mathbb{R}^{E \times E} \) is the inhibition mask matrix, and \( \gamma \) is the inhibition strength coefficient. This mechanism adjusts the membrane potential but does not alter the form of the input current \( I[t] \) calculation.
    \item As shown in equation \ref{equ:refractory1}, the refractory period mechanism affects membrane potential reset and spike emission. Here, \( R[t] \in \mathbb{R}^{B \times E} \) is the refractory period counter. When a neuron spikes, the membrane potential is reset to \( V[t] \leftarrow 0 \) and \( R[t] \leftarrow 2 \), but this does not involve the calculation of the input current.
\end{itemize}

Therefore, the input current \( I_i[t] \) is always composed of the weighted sum of the spike outputs from the presynaptic neurons, consistent with Definition \ref{def: spike-driven}. In summary, the routing and expert-selection processes in k-WTA Router exhibit spike-driven characteristics.
\end{proof}

\subsection{Experiment Details}
\label{suppl:experiment details}

\subsubsection{Datasets}
\label{suppl:datasets and task description}

\textbf{ImageNet-1K.} We use ImageNet-1K to validate the generalizability of the k-WTA Router on visual tasks. ImageNet \citep{deng2009imagenet} contains around 1,3 millon 1,000-class images for training and 50,000 images for validation. The input size of our experiments on ImageNet is set to the default \(224\times224\).

\textbf{GLUE.} We applied the standard GLUE benchmark \citep{wang2018glue}, using SpikeLM \citep{xing2024spikelm} as the baseline, to evaluate the performance of the k-WTA Router on natural language understanding tasks. We follow previous works and use the 8 subsets, including CoLA, STS-B, MRPC, RTE, QQP, MNLI and QNLI for classification or regression in different scenes. For evaluation, we follow BERT \citep{devlin2019bert} and report F1 scores for QQP and MRPC datasets; Spearman correlations for the STS-B dataset; and accuracy scores for other datasets. 

\textbf{CMU-MOSI.} CMU-MOSI \citep{zadeh2016mosi} is a multimodal sentiment analysis dataset comprising audio, video, and text modalities. It comprises 2,199 monologue video samples, with 1,284 used for training, 229 for validation, and 686 for testing. Acoustic and visual features are sampled at 12.5Hz and 15Hz.

\textbf{CMU-MOSEI.} CMU-MOSEI \citep{zadeh2018multimodal} utilizing the same modalities as CMU-MOSI. It includes 22,856 YouTube movie review clips, with 16,326 for training, 1,871 for validation, and 4,659 for testing. Acoustic and visual features are sampled at 20Hz and 15Hz. Sentiment labels on both datasets range from -3 (highly negative) to 3 (highly positive).

\textbf{UrbanSound8k-AV.} The UrbanSound8K-AV dataset \citep{guo2023transformer} merges the UrbanSound8K audio dataset \citep{salamon2014dataset} with its corresponding images, comprising 8732 audiovisual samples. Each sample consists of a high-resolution color image and a 4-second audio signal. We follow Guo et al. \citep{guo2023transformer} by randomly dividing the dataset into training and test sets in a 7:3 ratio.

The image classification (ImageNet-1K) experiments were conducted on 4 NVIDIA H100 GPUs, while the remaining experiments were conducted on 8 NVIDIA RTX 4090 GPUs. To evaluate the generalizability of the k-WTA Router, we conducted experiments by replacing the FFN layer in the baseline model with the k-WTA Router. For assessing robustness under missing modalities, we conducted experiments using two experimental settings (refer to Sec.\ref{sec: experiments settings}) on CMU-MOSI and CMU-MOSEI, respectively, while only experiment setting I was employed for experiments on UrbanSound8k-AV. The comparison between CMU-MOSI and CMU-MOSEI includes all 8 baseline methods, while the comparison for UrbanSound8k-AV only includes the SNN baseline method. For ease of comparison, we use F1 scores and AUC values as evaluation metrics for each method on CMU-MOSI and CMU-MOSEI; we use accuracy as the evaluation metric for each method on UrbanSound8k-AV.

\subsubsection{Model Details}
\label{suppl: model details}

In this section, we supplement the model details not covered in the main text.

\textbf{Modality-Specific Encoders.} To eliminate the confusing effects arising from different encoder architectures, we employ a modality-specific encoder with the same architecture across all models, as variations in encoder design significantly impact final performance. 

Spiking visual encoder employs a dual-path hierarchical architecture, performing spike encoding separately on the channel dimension and feature dimension. Given visual data \(X_v \in \mathbb{R}^{B \times C \times F}\), where \(B\) is the batch size, \(C\) is the number of channels, and \(F\) is the feature dimension, the encoding process can be represented as: 
\begin{align}  
S_c &= \mathcal{SN}(W_c \operatorname{LN}(X_v^\top)), \\  
S_f &= \mathcal{SN}(W_f \operatorname{LN}(S_c^\top)),  
\end{align}  
where \(\mathcal{SN}\) denotes spiking neuron layer, \(W_c \in \mathbb{R}^{N \times C}\) denotes the channel projection weights, \(W_f \in \mathbb{R}^{N \times C}\) denotes the feature projection weights, and \(\text{LN}\) represents LayerNorm. The final output is \(\mathcal{E}_v=\{S_f\}^T \in \mathbb{R}^{T \times B \times N \times C}\).

Spiking audio encoder employs temporal slicing and spiking convolution, enhancing local feature extraction capabilities by segmenting long-duration signals and introducing spatio-temporal convolution. For the audio signal \(X_a \in \mathbb{R}^{B \times S \times E}\), temporal segmentation is first performed:
\begin{align}
P &= \lceil S/L_P \rceil, \\
\tilde X_a &= \text{Padding}(X_a, P \times L_P-S), \\
X_{patch}&=\text{Unfold}(\tilde X_a, \text{kernel size}, \text{stride}).
\end{align}

Subsequently, temporal feature extraction and pulse coding are performed:
\begin{align}
S_P &= \mathcal{SN}(W_P \cdot \text{LN}(X_{patch})), \\
S_a &= \mathcal{SN}(\text{Conv}(S_P)),
\end{align}
where \(L_P\) denotes the block length, \(\text{Unfold}\) represents the sliding window operation, and \(\text{Conv}\) employs group convolutions to enhance feature diversity.

Spiking text encoder processes short, medium, and long-term dependencies in text sequences through parallel processing by LIF neurons with varying time constants. For text embedding \(X_t \in \mathbb{R}^{B \times L \times d}\), the encoding process is as follows:
\begin{align}
& S_f=\mathcal{SN}_{\tau=1.5}(W_f \cdot X_t), \\
& S_m=\mathcal{SN}_{\tau=2.0}(W_m \cdot X_t), \\
& S_s=\mathcal{SN}_{\tau=3.0}(W_s \cdot X_t),
\end{align}
where \(S_f\), \(S_m\), and \(S_s\) correspond to branches with different time constants. Feature fusion is achieved through concatenation and projection to obtain \(S_{fused}\), and finally, feature information is integrated using residual connection:
\begin{align}
& S_{res}=\mathcal{SN}(W_{res} \cdot X_t),\\
& S_{t}=\text{LN}(S_{fused}+S_{res}).
\end{align}

For SNNs methods, we uniformly employ the aforementioned encoding process to construct modality-specific spike encoders; for ANNs methods, we utilize pulse encoders with their architecturally equivalent counterparts within ANNs.

\textbf{Multi-head Spiking Self Attention.} We extend the Spike-Driven Self-Attention (SDSA) proposed in \citep{yao2023spike} to a multi-head architecture. Specifically, the floating-point matrices \(Q\), \(K\), and \(V\) are computed by three learnable linear matrices, followed by pulse neurons that transform \(Q\), \(K\), and \(V\) into spiking tensors \(Q_S\), \(K_S\), and \(V_S\). SDSA is expressed as:
\begin{equation}
    \text{SDSA}(Q,K,V)=\mathcal{SN}(\text{SUM}_c(Q_S \otimes K_S))\otimes V_S,
\end{equation}
where $\otimes$ denotes the Hadamard product, and $\text{SUM}_c$ represents the sum of each column. The Hadamard product between spiking tensors corresponds to a mask operation \citep{yao2023spike}.

In practice, we reshape \(Q\), \(K\), and \(V\) into multi-head formats. We then divide \(Q\), \(K\), and \(V\) into \(H\) segments and perform parallel H SDSA operations, referred to as H-head SDSA. The Multi-head Spiking Self-Attention (MSSA) in SpikeMoE is represented as follows:
\begin{align}
Q = (q_1,&\dots,q_H),\\
K = (k_1,&\dots,k_H),\\
V = (v_1,&\dots,v_H),\\
\tilde{\text{M}} = \bigl[\text{SDSA}_1(q_1,k_1,v_1);
                 & \dots;
                 \text{SDSA}_1(q_1,k_1,v_1)\bigr],\\[2pt]
\text{MSSA}(Q,K,V) &= \text{Linear}\bigl(\tilde{\text{M}}).
\end{align}

\textbf{Spiking Expert.} Each of the \(K\) experts is a spike-based MLP. Given an input \(X \in \mathbb{R}^{T \times B \times N \times C}\), the \(k\)-th expert computes:
\begin{align}
& \text{E}=\mathcal{SN}(\text{Linear}(\text{BN}(X))), \\
& \text{Expert}_k(\text{E})=\mathcal{SN}(\text{Linear}(\text{BN}(\text{E}))),
\end{align}
where \(\text{BN}\) is BatchNorm.
 
\subsubsection{Baseline Methods}
\label{suppl: baseline methods}

\textbf{ANN Baselines.} \textbf{ShaSpec} \citep{wang2023multi} employs multiple loss functions to learn both modality-shared and modality-specific information, and utilizes the shared information to estimate missing modalities. \textbf{TF} \citep{zadeh2017tensor} extracts modality interactions through multimodal sub-networks and a tensor fusion layer. \textbf{mmFormer} \citep{zhang2022mmformer}, a typical multimodal Transformer, features multi-head self-attention mechanisms. \textbf{FuseMoE} \citep{han2024fusemoe} respects missing modalities and jointly learns from both single and missing modality combinations of data inputs. In experiments, FuseMoE utilizes a softmax gate.

\textbf{SNN Baselines.} \textbf{Spikformer} \citep{zhou2022spikformer} pioneers the integration of Transformer architectures into SNNs by proposing Spiking Self-Attention. It significantly enhances the representation capacity of SNNs for visual feature extraction while maintaining biological interpretability. \textbf{Spike-driven Transformer} \citep{yao2023spike} introduces a purely event-driven attention paradigm. It effectively replaces all dense matrix multiplications in vanilla Transformers with sparse addition operations based on spike masks. \textbf{Spikingformer} \citep{zhou2023spikingformer} further improves SNN-based Transformers by optimizing spike-driven residual learning and feature mapping mechanisms. \textbf{SpikeLM} \citep{xing2024spikelm} is an advanced spiking language model specifically designed for natural language processing (NLP) tasks. By adapting spike representations for text sequences, it demonstrates the effectiveness and scalability of SNNs on standard language understanding benchmarks such as GLUE. \citep{zhou2022spikformer} .\textbf{WeightAttention} \citep{liu2022event} dynamically assigns weights to measure the importance of visual and auditory modalities, thereby effectively integrating information from these two modalities. \textbf{SCA} \citep{guo2023transformer} integrates image and audio sub-modules and combines relative positional bias (RPB) to enhance the ability to capture information from different positions. \textbf{S-CMRL} \citep{he2025enhancing} employs spatiotemporal spike attention to extract complementary features across modalities and combines cross-modal residual learning to enhance feature integration. Additionally, S-CMRL introduces a semantic alignment optimization mechanism to align cross-modal features within a shared semantic space, thereby improving their consistency and robustness.

\begin{figure*}[htbp]
    \centering    
    \includegraphics[width=0.9\textwidth]{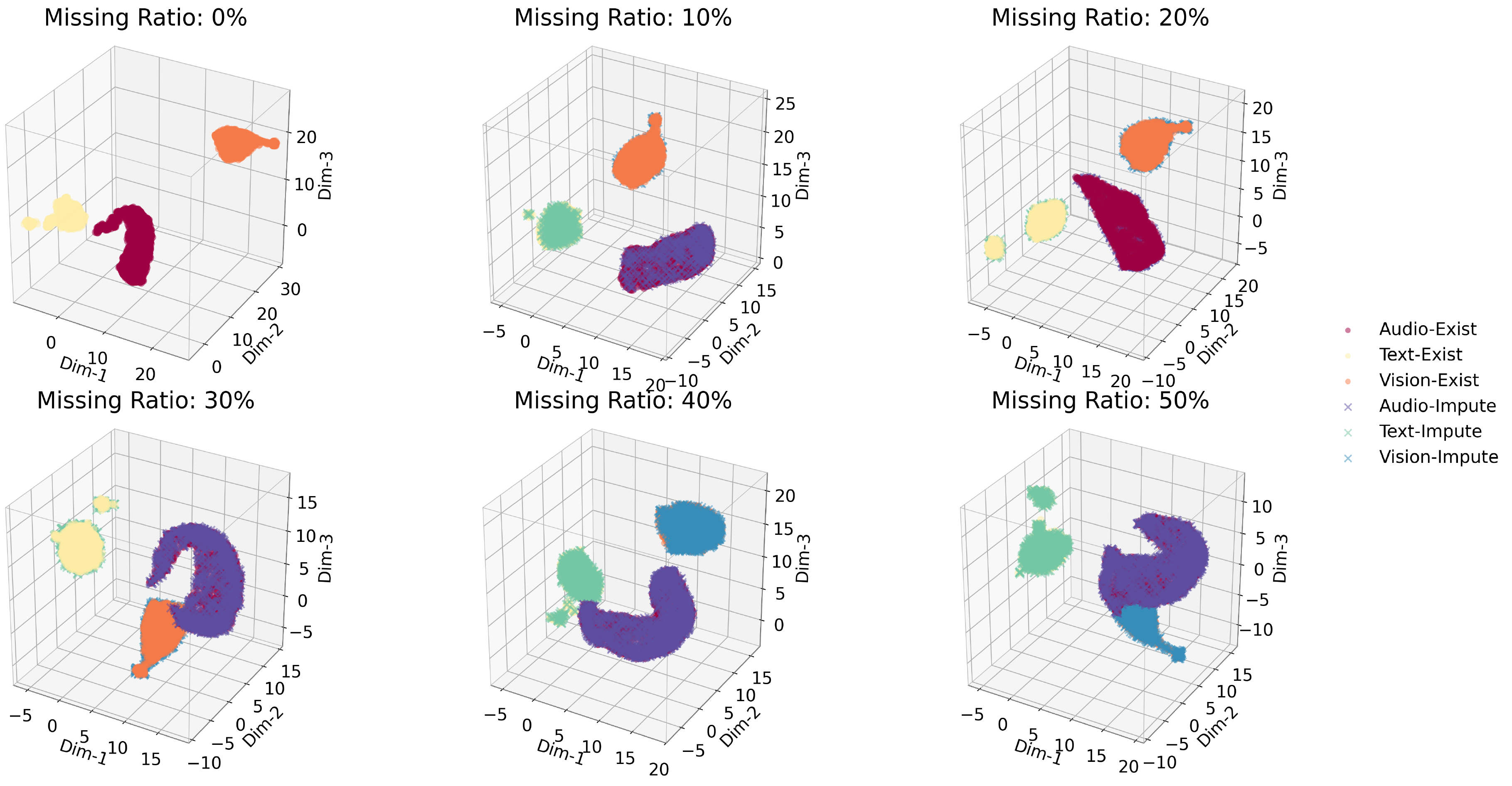}  
    \caption{UMAP visualization of multi-modal embedding space: CMU-MOSEI.}
    \label{fig:umap_mosei}
\end{figure*}


\subsubsection{Complexity and Energy Consumption Analysis}
\label{suppl:complexity analysis}

We compare the complexity of our proposed method and baseline models in terms of operations (OPs) and parameters, along with their AUC values on the CMU-MOSI dataset at a missing ratio of 50\%. In SNNs, OPs refer to SOPs, while in ANNs they refer to FLOPs. SOPs are computed as follows:
\begin{equation}
    \text{SOPs}(l)=fr \times T \times \text{FLOPs}(l),
\end{equation}
where \(l\) denotes the block/layer within the Transformer architecture, \(fr\) represents the emission rate of the input pulse sequence for the block/layer, and \(T\) is the simulated time step for spike neuron. \(\text{FLOPs}(l)\) refers to the number of floating-point operations for \(l\), specifically the count of multiply-accumulate (MAC) operations. SOPs denote the number of spike-based accumulate (AC) operations.

For all theoretical energy consumption calculations in this paper, we adopt the theoretical energy consumption estimation method proposed by \citep{yao2023spike}. The results demonstrate that SpikeMoE achieves the optimal balance between complexity and performance. Detailed data for the complexity analysis are provided in Table \ref{tab:complexity}. The results demonstrate that ShaSpec is the simplest model, yet its task performance is not comparable. In contrast, mmformer and S-CMRL are highly complex in terms of OPs and parameters, but their performance is not optimal. FuseMoE demonstrates performance comparable to our proposed method, yet it requires significantly more OPs and parameters. SpikeMoE is the optimal model that best balances complexity and performance.

\begin{table*}[t]
\centering
\caption{Complexity and Energy Consumption analysis}
\label{tab:complexity}
\resizebox{\textwidth}{!}{
\begin{tabular}{cccccccccc}
\toprule
\multicolumn{1}{c}{} &
\multicolumn{4}{c}{\textbf{ANNs}} &
\multicolumn{3}{c}{\textbf{SNNs}} &
\multicolumn{1}{c}{\textbf{Ours}} \\
\cmidrule(lr){2-5}\cmidrule(lr){6-8}\cmidrule(lr){9-9}
\textbf{Metric} & \textbf{ShaSpec} & \textbf{TF} & \textbf{mmFormer} & \textbf{FuseMoE} &
\textbf{WeightAttention} & \textbf{SCA} & \textbf{S-CMRL} &
\textbf{SpikeMoE} \\
\midrule
Ops (M) & 21.73 & 48.42 & 1577.63 & 199.8 & 47.07 & 32.57 & 289.60 & \textbf{45.22} \\
Params &  2,177,406 & 4,211,331 & 11,373,315 & 243,543,180 & 131,585 & 264,192 & 2,685,729 & \textbf{2,472,370} \\
AUC & 55.60$\pm$0.12 & 60.49$\pm$0.80 & 61.87$\pm$0.67 & 67.28$\pm$0.28 &
65.81$\pm$0.05 & 64.49$\pm$0.43 & 65.71$\pm$0.43 &
\textbf{67.30$\pm$0.71} \\
Energy (mJ) & 20.13 & 25.19 & 40.39 & 11.71 & 10.38 & 16.95 & 14.67 & \textbf{5.29} \\
\bottomrule
\end{tabular}}
\end{table*}

\subsubsection{Visualization of Experiment}
\label{suppl:visualization of experiment}

In this section, we provide visual evidence to further support the effectiveness of our proposed missing modality modeling approach. We present the UMAP projection of multimodal embeddings on CMU-MOSEI in Figure \ref{fig:umap_mosei}. The visualization clearly demonstrates that the interpolated modalities form distributions consistent with their original counterparts. Combined with the results in Figure \ref{fig:umap}, this indicates that our proposed two-stage missing-modality modeling strategy preserves semantic alignment across modalities, proving its effectiveness in the missing-modality filling process.

\subsubsection{Implementation Details and Hardware Setting}
\label{suppl: implementation details and hardware setting}

For the spiking neuron model, LIF neurons are configured with an initial membrane potential of 0 and a firing threshold of 0.5. To facilitate efficient backpropagation, a sigmoid function with parameter \(\alpha=4\) is employed as the surrogate gradient function, expressed as \(Sigmoid(x)=1/(1+\text{exp}(-\alpha x))\). Key hyperparameters are listed in Table \ref{tab:hparam}. SpikeMoE is optimized through the minimization of the expected risk based on the loss function \(L_{balance}\). The load-balance loss \(\mathcal{L}_{\text{balance}}\) encourages the equitable utilization of experts, preventing some from being overused while others are neglected. By integrating these losses, we achieve the overall objective:
\begin{align}
& \mathcal{L}_{\text{balance}}=\text{Var}(p)+\alpha(\text{log} \ E-\text{Entropy}(\mathbf{p})), \\
& \mathcal{L}_{\text{total}}=\mathcal{L}_{\text{task}}+\lambda \mathcal{L}_{\text{balance}},
\end{align}
where  \(\lambda\) is the coefficient of \(\mathcal{L}_{\text{balance}}\). \(\mathcal{L}_{\text{task}}\) is the cross-entropy loss.

\subsubsection{Discussion of Sensitivity Analysis}
\label{suppl: discussion of sensitivity analysis}

To thoroughly evaluate the impact of hyperparameters on model performance, we conduct a sensitivity analysis on the CMU-MOSEI dataset (missing ratio=50\%). We analyze the number of experts, number of expert layers, WTA-timesteps, and inhibition strength (Figure \ref{fig:sensitivity}). The detailed discussion is as follows:

\textbf{Model Capacity and Depth}: As the number of experts increases from 4 to 8 and the expert layers from 2 to 5, both AUC and F1 metrics show a steady upward trend. This indicates that expanding the expert pool and deepening the MoE structure effectively enhance the network's ability to fit complex data distributions. Consequently, the model learns more fine-grained and hierarchically rich specialized features. This also demonstrates the promising scalability of SpikeMoE.

\textbf{Temporal Dynamics}: In the temporal dimension, the model achieves optimal performance when the WTA-timesteps are set to 14. If the time steps are too short, neurons cannot sufficiently accumulate membrane potential, depriving routing decisions of adequate temporal dynamics. Conversely, performance slightly declines when the time steps exceed 14. This is likely because excessively long integration introduces temporal noise without providing additional representation benefits.

\textbf{Competition-Inhibition Balance}: For the inhibition strength—a core parameter of the spiking gating network—the model peaks in performance at 0.3. This validates the importance of dynamically balanced selection. A moderate inhibition strength (0.3) successfully prevents expert collapse and ensures sparse activation. However, overly weak inhibition (< 0.3) reduces competition among experts, degrading routing sparsity. Conversely, excessively strong inhibition (> 0.3) causes over-suppression, making it difficult for certain expert neurons to activate and thereby limiting collaborative feature processing across multiple experts.

\begin{table}[H]
\centering
\vspace{-5mm}
\caption{Hyperparameter setting}
\label{tab: hyperparameter setting}
\label{tab:hparam}
\renewcommand{\arraystretch}{1.3}
\begin{tabular}{>{\centering\arraybackslash}p{0.3\linewidth} | >{\centering\arraybackslash}p{0.6\linewidth}}
\toprule
\textbf{Parameter} & \textbf{Value} \\
\midrule
Number of Expert & 4 for ImageNet-1K and GLUE, 6 for CMU-MOSI and CMU-MOSEI, 4 for UrbanSound8k-AV\\
Top-K Expert & 2\\
Timesteps \(T\) & 4 for ImageNet-1K, GLUE and UrbanSound8K-AV; 6 for CMU-MOSI and CMU-MOSEI\\
Learning Rate & 5e-5\\
Hidden Size & 128\\
MoE layers & 2\\
K for Sampling & 10 \\
wta-timesteps \(t\) & 14 \\
Refractory Period & 2 \\
Inhibition Strength & 0.3 \\
Training Epoch & 300\\
Random Seed & 42, 43, 44\\
Dropout & 0.3\\
\bottomrule
\end{tabular}
\end{table}

\subsection{Results on CMU-MOSEI}
\label{supp:results on CMU-MOSEI}

We conducted experiments using experimental setup I and II on the CMU-MOSEI dataset, with results shown in Table \ref{tab:Experiment Setting I: CMU-MOSEI} and \ref{tab:Experiment Setting II: CMU-MOSEI}.

\begin{table*}[t]
\centering
\setlength{\tabcolsep}{1.4pt}
\renewcommand{\arraystretch}{1.1}
\setlength{\fboxsep}{1.0pt}
\caption{Experiment Setting I: Main results on CMU-MOSEI. Top-3 results are highlighted as \smash{\colorbox{firstcolor}{first}}, \smash{\colorbox{secondcolor}{second}} and \smash{\colorbox{thirdcolor}{third}}.}
\label{tab:Experiment Setting I: CMU-MOSEI}
\resizebox{\textwidth}{!}{%
\begin{tabular}{cccccccccc}
\toprule
\multirow{3}{*}{\shortstack[c]{\textbf{Missing}\\\textbf{Ratio}}} &
\multirow{3}{*}{\textbf{Metric}} &
\multicolumn{4}{c}{\textbf{ANNs}} &
\multicolumn{3}{c}{\textbf{SNNs}} &
\multicolumn{1}{c}{\textbf{Ours}} \\
\cmidrule(lr){3-6}\cmidrule(lr){7-9}\cmidrule(lr){10-10}
& &
\shortstack[c]{\textbf{ShaSpec}\\\citep{wang2023multi}} &
\shortstack[c]{\textbf{TF}\\\citep{zadeh2017tensor}} &
\shortstack[c]{\textbf{mmFormer}\\\citep{zhang2022mmformer}} &
\shortstack[c]{\textbf{FuseMoE}\\\citep{han2024fusemoe}} &
\shortstack[c]{\textbf{Weight Attention}\\\citep{liu2022event}} &
\shortstack[c]{\textbf{SCA}\\\citep{guo2023transformer}} &
\shortstack[c]{\textbf{S-CMRL}\\\citep{he2025enhancing}} &
\textbf{SpikeMoE} \\
\midrule
\multirow{2}{*}{0\%}
 & F1  & 47.48$\pm$0.27 & 38.89$\pm$0.54 & 56.75$\pm$0.47 & 57.04$\pm$0.15 & \smash{\colorbox{secondcolor}{59.40$\pm$0.42}} & \smash{\colorbox{thirdcolor}{59.36$\pm$0.06}} & 58.09$\pm$0.19 & \smash{\colorbox{firstcolor}{59.80$\pm$0.17}} \\
 & AUC & 67.93$\pm$0.56 & 64.26$\pm$0.61 & \smash{\colorbox{thirdcolor}{76.42$\pm$0.24}} & \smash{\colorbox{secondcolor}{78.54$\pm$0.42}} & 73.59$\pm$0.08 & 74.41$\pm$0.13 & 72.63$\pm$0.21 & \smash{\colorbox{firstcolor}{78.63$\pm$0.53}} \\
\midrule
\multirow{2}{*}{10\%}
 & F1  & 47.41$\pm$0.18 & 34.94$\pm$1.10 & 55.37$\pm$0.47 & 55.90$\pm$0.55 & 54.97$\pm$0.35 & \smash{\colorbox{secondcolor}{56.81$\pm$0.35}} & \smash{\colorbox{thirdcolor}{56.23$\pm$0.32}} & \smash{\colorbox{firstcolor}{56.83$\pm$0.23}} \\
 & AUC & 69.26$\pm$0.22 & 61.05$\pm$0.34 & \smash{\colorbox{thirdcolor}{75.36$\pm$0.53}} & \smash{\colorbox{secondcolor}{76.38$\pm$0.39}} & 72.18$\pm$0.34 & 73.12$\pm$0.23 & 73.17$\pm$0.09 & \smash{\colorbox{firstcolor}{76.47$\pm$0.49}} \\
\midrule
\multirow{2}{*}{20\%}
 & F1  & 46.54$\pm$1.04 & 35.37$\pm$0.94 & 52.91$\pm$0.75 & \smash{\colorbox{firstcolor}{54.44$\pm$0.27}} & 53.14$\pm$0.17 & \smash{\colorbox{thirdcolor}{54.21$\pm$0.11}} & 52.84$\pm$0.45 & \smash{\colorbox{secondcolor}{54.35$\pm$0.57}} \\
 & AUC & 68.05$\pm$0.56 & 59.54$\pm$0.22 & \smash{\colorbox{thirdcolor}{73.02$\pm$0.65}} & \smash{\colorbox{secondcolor}{74.32$\pm$0.16}} & 69.92$\pm$0.12 & 70.32$\pm$0.54 & 69.95$\pm$0.02 & \smash{\colorbox{firstcolor}{74.92$\pm$0.86}} \\
\midrule
\multirow{2}{*}{30\%}
 & F1  & 45.05$\pm$0.90 & 35.36$\pm$0.84 & 49.71$\pm$0.30 & \smash{\colorbox{secondcolor}{52.27$\pm$0.35}} & 51.60$\pm$0.55 & \smash{\colorbox{thirdcolor}{51.78$\pm$0.31}} & 50.51$\pm$0.03 & \smash{\colorbox{firstcolor}{54.22$\pm$0.49}} \\
 & AUC & 66.89$\pm$0.31 & 60.23$\pm$0.31 & \smash{\colorbox{thirdcolor}{71.28$\pm$0.10}} & \smash{\colorbox{secondcolor}{72.22$\pm$0.52}} & 68.54$\pm$0.07 & 68.14$\pm$0.50 & 67.46$\pm$0.33 & \smash{\colorbox{firstcolor}{73.66$\pm$0.41}} \\
\midrule
\multirow{2}{*}{40\%}
 & F1  & 43.49$\pm$0.95 & 32.23$\pm$0.61 & 48.06$\pm$0.54 & \smash{\colorbox{firstcolor}{49.32$\pm$0.17}} & \smash{\colorbox{thirdcolor}{48.30$\pm$0.03}} & 47.90$\pm$0.39 & 48.14$\pm$0.41 & \smash{\colorbox{secondcolor}{48.89$\pm$0.62}} \\
 & AUC & 65.32$\pm$0.59 & 60.09$\pm$0.35 & \smash{\colorbox{thirdcolor}{69.31$\pm$0.10}} & \smash{\colorbox{secondcolor}{69.85$\pm$0.11}} & 67.40$\pm$0.47 & 63.98$\pm$0.12 & 66.75$\pm$0.36 & \smash{\colorbox{firstcolor}{69.92$\pm$0.68}} \\
\midrule
\multirow{2}{*}{50\%}
 & F1  & 42.75$\pm$1.22 & 33.01$\pm$0.51 & \smash{\colorbox{thirdcolor}{47.39$\pm$0.16}} & 43.08$\pm$0.47 & 46.50$\pm$0.54 & 46.49$\pm$0.26 & \smash{\colorbox{secondcolor}{47.44$\pm$0.24}} & \smash{\colorbox{firstcolor}{49.82$\pm$0.74}} \\
 & AUC & 62.32$\pm$0.42 & 58.20$\pm$0.17 & \smash{\colorbox{thirdcolor}{66.61$\pm$0.22}} & \smash{\colorbox{secondcolor}{67.63$\pm$0.55}} & 63.84$\pm$0.39 & 63.06$\pm$0.05 & 64.09$\pm$0.07 & \smash{\colorbox{firstcolor}{68.80$\pm$0.20}} \\
\bottomrule
\end{tabular}%
}
\vspace{-6mm}
\end{table*}

\begin{table*}[t]
\centering
\setlength{\tabcolsep}{1.4pt}
\renewcommand{\arraystretch}{1.1}
\setlength{\fboxsep}{1.0pt}
\caption{Experiment Setting II: Main results on CMU-MOSEI. Top-3 results are highlighted as \smash{\colorbox{firstcolor}{first}}, \smash{\colorbox{secondcolor}{second}} and \smash{\colorbox{thirdcolor}{third}}.}
\label{tab:Experiment Setting II: CMU-MOSEI}
\resizebox{\textwidth}{!}{%
\begin{tabular}{cccccccccccc}
\toprule
\multicolumn{3}{c}{\textbf{Testing Modality}} &
\multirow{2}{*}{\textbf{Metric}} &
\multicolumn{4}{c}{\textbf{ANNs}} &
\multicolumn{3}{c}{\textbf{SNNs}} &
\multicolumn{1}{c}{\textbf{Ours}} \\
\cmidrule(lr){1-3}\cmidrule(lr){5-8}\cmidrule(lr){9-11}\cmidrule{12-12}
\textbf{Video} & \textbf{Text} & \textbf{Audio} & &
\shortstack[c]{\textbf{ShaSpec}\\\citep{wang2023multi}} &
\shortstack[c]{\textbf{TF}\\\citep{zadeh2017tensor}} &
\shortstack[c]{\textbf{mmFormer}\\\citep{zhang2022mmformer}} &
\shortstack[c]{\textbf{FuseMoE}\\\citep{han2024fusemoe}} &
\shortstack[c]{\textbf{Weight Attention}\\\citep{liu2022event}} &
\shortstack[c]{\textbf{SCA}\\\citep{guo2023transformer}} &
\shortstack[c]{\textbf{S-CMRL}\\\citep{he2025enhancing}} &
\textbf{SpikeMoE} \\
\midrule
\checkmark & \checkmark & & F1 &
40.75$\pm$0.14 & 31.38$\pm$0.61 &
\smash{\colorbox{secondcolor}{56.02$\pm$0.56}} &
\smash{\colorbox{thirdcolor}{55.85$\pm$0.69}} &
33.34$\pm$0.28 & 52.80$\pm$0.76 & 55.62$\pm$0.34 &
\smash{\colorbox{firstcolor}{56.25$\pm$0.32}} \\
& & & AUC &
64.47$\pm$0.10 & 56.98$\pm$0.20 &
\smash{\colorbox{secondcolor}{76.60$\pm$0.39}} &
\smash{\colorbox{thirdcolor}{75.81$\pm$0.82}} &
65.91$\pm$0.45 & 71.60$\pm$0.04 & 73.21$\pm$0.45 &
\smash{\colorbox{firstcolor}{77.09$\pm$0.44}} \\
\midrule
\checkmark & & \checkmark & F1 &
\smash{\colorbox{thirdcolor}{38.15$\pm$0.92}} &
31.73$\pm$0.16 & 37.12$\pm$0.66 &
\smash{\colorbox{secondcolor}{49.30$\pm$0.67}} &
22.00$\pm$0.49 & 29.01$\pm$0.04 & 22.10$\pm$0.20 &
\smash{\colorbox{firstcolor}{49.50$\pm$0.36}} \\
& & & AUC &
\smash{\colorbox{firstcolor}{58.41$\pm$0.10}} &
\smash{\colorbox{firstcolor}{58.41$\pm$0.17}} &
\smash{\colorbox{thirdcolor}{56.72$\pm$0.84}} &
56.32$\pm$0.62 & 50.91$\pm$0.57 & 51.26$\pm$0.25 &
50.30$\pm$0.06 &
\smash{\colorbox{secondcolor}{57.08$\pm$0.24}} \\
\midrule
& \checkmark & \checkmark & F1 &
40.82$\pm$0.12 & 31.86$\pm$0.23 &
\smash{\colorbox{thirdcolor}{56.06$\pm$0.29}} &
\smash{\colorbox{secondcolor}{57.02$\pm$0.60}} &
\smash{\colorbox{thirdcolor}{56.06$\pm$0.48}} &
33.90$\pm$0.21 & 21.93$\pm$0.83 &
\smash{\colorbox{firstcolor}{58.53$\pm$0.02}} \\
& & & AUC &
64.46$\pm$0.14 & 58.64$\pm$0.13 &
\smash{\colorbox{secondcolor}{73.91$\pm$0.04}} &
\smash{\colorbox{thirdcolor}{73.42$\pm$0.60}} &
73.31$\pm$0.10 & 56.54$\pm$0.17 & 50.29$\pm$0.32 &
\smash{\colorbox{firstcolor}{75.47$\pm$0.19}} \\
\midrule
\checkmark & & & F1 &
40.37$\pm$0.46 & 31.59$\pm$0.23 &
\smash{\colorbox{secondcolor}{55.07$\pm$0.35}} &
\smash{\colorbox{thirdcolor}{54.02$\pm$2.88}} &
32.65$\pm$0.44 & 50.80$\pm$0.61 & 53.41$\pm$0.29 &
\smash{\colorbox{firstcolor}{55.09$\pm$0.42}} \\
& & & AUC &
63.52$\pm$0.19 & 58.71$\pm$0.35 &
\smash{\colorbox{secondcolor}{75.82$\pm$0.28}} &
\smash{\colorbox{thirdcolor}{74.84$\pm$0.14}} &
64.96$\pm$0.59 & 72.06$\pm$0.39 & 70.62$\pm$0.02 &
\smash{\colorbox{firstcolor}{76.03$\pm$0.39}} \\
\midrule
& \checkmark & & F1 &
42.16$\pm$0.22 & 30.84$\pm$0.26 &
\smash{\colorbox{firstcolor}{55.89$\pm$0.37}} &
\smash{\colorbox{secondcolor}{55.87$\pm$0.68}} &
33.20$\pm$0.68 & 51.04$\pm$0.03 & 52.32$\pm$1.25 &
\smash{\colorbox{thirdcolor}{55.62$\pm$0.25}} \\
& & & AUC &
64.40$\pm$0.24 & 58.89$\pm$0.40 &
\smash{\colorbox{firstcolor}{74.83$\pm$0.28}} &
\smash{\colorbox{thirdcolor}{73.83$\pm$0.05}} &
65.02$\pm$0.07 & 71.02$\pm$0.06 & 69.98$\pm$0.33 &
\smash{\colorbox{secondcolor}{74.23$\pm$0.22}} \\
\midrule
& & \checkmark & F1 &
42.45$\pm$0.85 & 30.71$\pm$0.71 &
\smash{\colorbox{thirdcolor}{55.06$\pm$0.29}} &
\smash{\colorbox{secondcolor}{55.64$\pm$1.65}} &
49.06$\pm$0.65 & 52.01$\pm$0.51 & 22.93$\pm$0.35 &
\smash{\colorbox{firstcolor}{55.93$\pm$0.47}} \\
& & & AUC &
65.16$\pm$0.04 & 58.32$\pm$0.04 &
\smash{\colorbox{thirdcolor}{75.85$\pm$0.09}} &
\smash{\colorbox{secondcolor}{76.14$\pm$0.35}} &
70.31$\pm$0.51 & 70.16$\pm$0.34 & 50.21$\pm$0.14 &
\smash{\colorbox{firstcolor}{76.84$\pm$0.15}} \\
\bottomrule
\end{tabular}%
}
\vspace{-4mm}
\end{table*}

\subsection{Results on UrbanSound8K-AV}
\label{supp:results on UrbanSound8k-AV}

\subsubsection{Primary Results}
\label{supp:primary results on UrbanSound8k-AV}

We conducted experiments using experimental setup I on the UrbanSound8k-AV dataset, with results shown in the Figure \ref{fig:results and ablation urbansound}. The experimental results demonstrate that SpikeMoE achieves state-of-the-art performance when compared against all SNN baseline methods.

\subsubsection{Ablation Study}
\label{supp:ablation study on UrbanSound9k-AV}

\begin{figure*}[htbp]
    \centering    
    \includegraphics[width=1.0\textwidth]{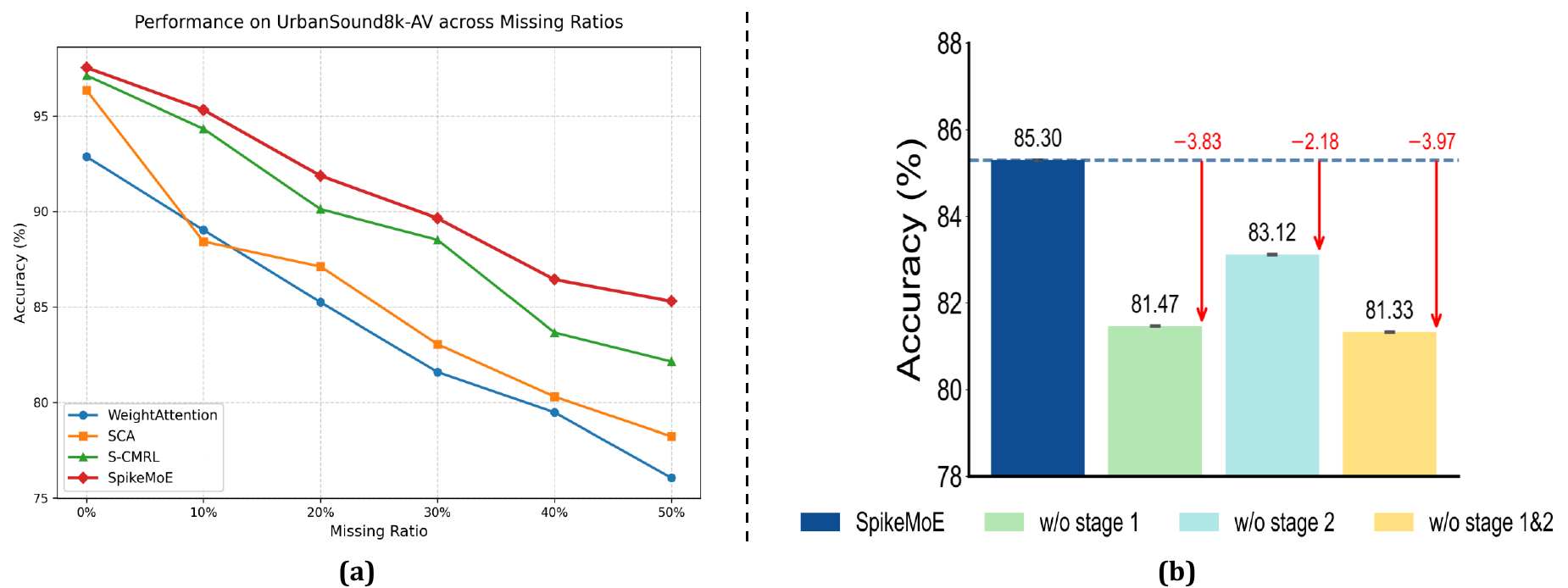}  
    \vspace{-5mm}
    \caption{(a) Performance on UrbanSound8k-AV across Missing Ratios; (b) Ablation study on UrbanSound8k-AV (Experiment Setting I, Missing Ratio 50\%).}
    \vspace{-3mm}
    \label{fig:results and ablation urbansound}
\end{figure*}

We conducted ablation studies on UrbanSound8k-AV as outlined in Fig.~\ref{fig:results and ablation urbansound}. We sequentially discarded the two stages of the missing modality modeling strategy and replaced missing modalities with zeros. When both stages were removed, Acc decreased by 3.97\% on UrbanSound8k-AV. Additionally, we observed the same conclusion on CMU-MOSI and CMU-MOSEI: removing stage 1 resulted in a greater decline than removing stage 2. This suggests that in missing modal modeling, the modal interactions captured by the modal pool may play a more critical role.

\subsection{Inference Latency and Throughput Analysis}
\label{suppl:inference latency and throughput analysis}

\begin{figure*}[htbp]
    \centering
    \includegraphics[width=1.0\textwidth]{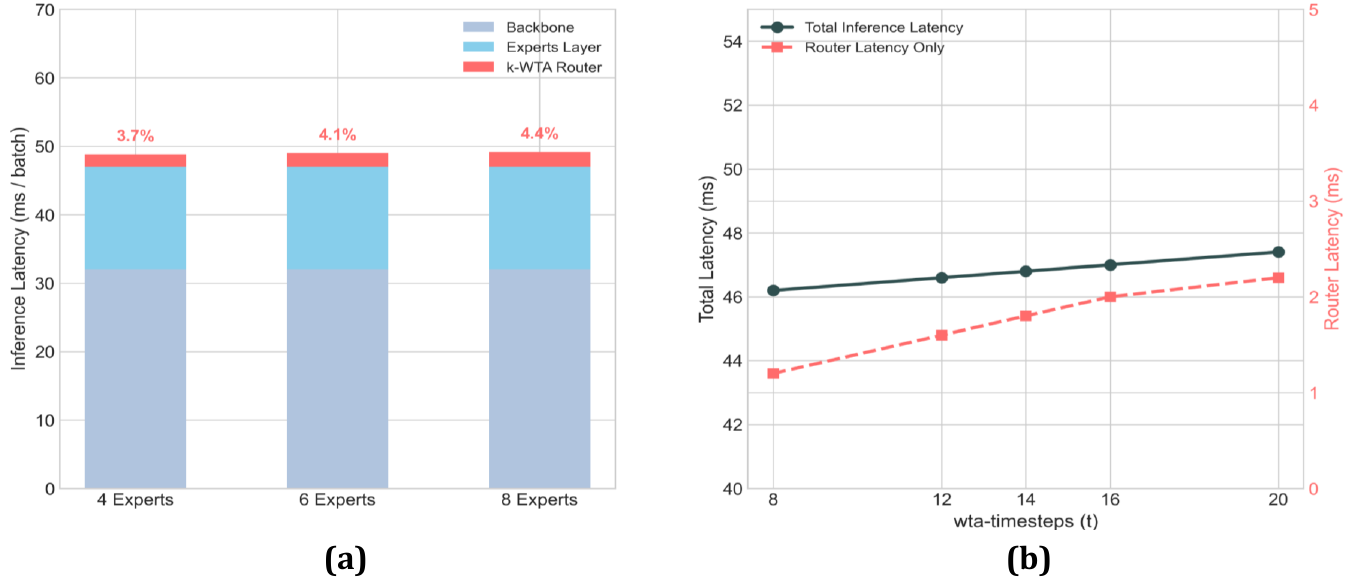}  
    \vspace{-5mm}
    \caption{(a) Latency Breakdown across Expert Counts; (b) Impact of wta-timesteps on Latency.}
    \vspace{-3mm}
    \label{fig:router latency}
\end{figure*}

We evaluated the inference latency of SpikeMoE on a single NVIDIA RTX 4090 GPU (Batch Size = 64). As shown in Fig.\ref{fig:router latency}, the left panel demonstrates that as the number of experts scales from 4 to 8, the computational overhead of the k-WTA router remains exceptionally low (accounting for only 3.7\% to 4.4\% of the total latency). Meanwhile, the backbone and expert layer latencies remain strictly constant due to the sparse Top-K activation. The right panel indicates that increasing the wta-timesteps (\(t\)) from 8 to 20 yields only a marginal increase in router execution time (\(\sim\)1 ms). Consequently, the total inference latency remains stable, confirming that our routing mechanism scales efficiently without introducing computational bottlenecks.

\subsection{Algorithm Workflow}
\label{suppl:algorithm workflow}

In this section, we present the algorithmic flow and pseudocode for the k-WTA Router (Algorithm \ref{alg:kWTARouter}).

\begin{algorithm}
\caption{k-WTA Router Algorithm}
\label{alg:kWTARouter}
\begin{algorithmic}[1]
\Require Input \(X \in \mathbb{R}^{T \times B \times N \times C}\), experts \(E\), wta-timesteps \(t_{max}\), time const \(\tau\), inhibit \(\gamma\), eps \(\varepsilon\)
\Ensure Routing weights \(w_{\text{expert}} \in \mathbb{R}^{B \times E}\), Selected experts \(\mathcal{E}_{\text{selected}}\)

\State \textbf{Initialize:} \(V \gets 0\), \(R \gets 0\), \(c \gets 0\), Weights \(L_{\text{inhibit}}\), Mask \(M_{\text{inhibit}}\)

\For{\(t = 1\) \textbf{to} \(t_{max}\)} \Comment{\textbf{Phase 1: Spiking \& Lateral Inhibition}}
    \State \(V[t] \gets V[t-1] \cdot (1 - \frac{1}{\tau}) + I[t]\)
    \State \(V_{\text{inhibit}}[t] \gets V[t] - S[t-1] \cdot (L_{\text{inhibit}} \odot M_{\text{inhibit}}) \cdot \gamma\)
    
    \For{\(b \in [1, B], e \in [1, E]\)}
        \If{\(V_{\text{inhibit}}[b,e,t] \geq V_{th}[t]\) \textbf{and} \(R[b,e,t] \leq 0\)}
            \State \(S[b,e,t] \gets 1\)
            \State \(V[b,e,t] \gets 0; \; R[b,e,t] \gets t_{\text{refra}}\) \Comment{Membrane reset \& refractory}
            \State \(c[b,e] \gets c[b,e] + 1\) \Comment{Accumulate spikes}
        \Else
            \State \(S[b,e,t] \gets 0\)
        \EndIf
    \EndFor
    
    \State \(R[t+1] \gets \max(0, R[t] - 1)\) \Comment{Update refractory counter}
\EndFor

\State \textbf{Expert Selection:} \Comment{\textbf{Phase 2: Routing Assignment}}
\State \(\mathcal{E}_{\text{selected}} \gets \text{TopK}_{i \in [E]}(c_i)\) \Comment{Select top-$K$ experts based on spike count}

\State \textbf{Compute Weights:} \(\forall b \in [1, B], \forall e \in [1, E]\)
\State \(w[b,e] \gets \begin{cases} 
      \frac{c[b,e] + \varepsilon}{\sum_{j \in \mathcal{E}_{\text{selected}}}(c[b,j] + \varepsilon)} & \text{if } e \in \mathcal{E}_{\text{selected}} \\
      0 & \text{otherwise}
   \end{cases}\)

\State \textbf{return} \(w_{\text{expert}}\), \(\mathcal{E}_{\text{selected}}\)
\end{algorithmic}
\end{algorithm}

\subsection{Limitations}
\label{appendix: limitations}

Limitations of this study include the need to test the performance of SpikeMoE when scaled up to larger models and more modalities, as well as limitations related to hardware deployment. These issues will be addressed in future research. In the future, we will evaluate SpikeMoE’s performance at larger scales and explore deploying SpikeMoE on hardware platforms such as Field-Programmable Gate Arrays (FPGAs) to assess its practical performance. In this process, we will optimize appropriate read-write data flows and memory access schemes to enhance the model’s inference speed.

\end{document}

%% file: math_commands.tex
\usepackage{amsmath,amsfonts,bm}

\def\eqref#1{equation~\ref{#1}}

\def\1{\bm{1}}

\DeclareMathAlphabet{\mathsfit}{\encodingdefault}{\sfdefault}{m}{sl}
\SetMathAlphabet{\mathsfit}{bold}{\encodingdefault}{\sfdefault}{bx}{n}

